\documentclass[twoside,11pt]{article}

\usepackage{jmlr2e}
\usepackage{amsmath}

\ShortHeadings{Confidence Intervals for Random Forests}{Wager, Hastie and Efron}
\firstpageno{1}

\usepackage{verbatim}
\usepackage{caption}
\usepackage{subcaption}
\usepackage{fancyvrb}
\usepackage[T1]{fontenc}

\usepackage{hyperref}

\usepackage{StefanTexCommands}

\graphicspath{{./}}

\newcommand{\hthRAW}{\htheta}  % base learner
\newcommand{\hthRF}{\htheta^{RF}} % random forest
\newcommand{\hthBAG}{\htheta^{B}} % bagged estimator
\newcommand{\hthBAGinf}{\htheta^{\infty}} % limiting bagged estimator

\newcommand{\vboot}{\hv}
\newcommand{\vn}{\tilde{v}^{(0)}}
\newcommand{\tvn}{\tilde{v}^{(+)}}

\newcommand{\hVJ}{\hV_{J}^B}
\newcommand{\hVJU}{\hV_{J-U}^B}
\newcommand{\hVJinf}{\hV_{J}^\infty}

\newcommand{\hVIJ}{\hV_{IJ}^B}
\newcommand{\hVIJU}{\hV_{IJ-U}^B}
\newcommand{\hVIJinf}{\hV_{IJ}^\infty}

\newcommand{\hsigma}{\hat{\sigma}}

\newcommand{\ooth}{{\Theta}}

\begin{document}

\title{Confidence Intervals for Random Forests: \\
The Jackknife and the Infinitesimal Jackknife}

\author{\name Stefan Wager \email swager@stanford.edu 
       \vspace{-1.5mm}
       \AND \name Trevor Hastie \email hastie@stanford.edu 
       \vspace{-1.5mm}
       \AND \name Bradley Efron \email brad@stat.stanford.edu
       \vspace{-1mm}
       \AND \addr Department of Statistics\\
       Stanford University\\
       Stanford, CA 94305, USA}

\maketitle

\begin{abstract}

We study the variability of predictions made by bagged learners and random forests, and show how to estimate standard errors for these methods. Our work builds on variance estimates for bagging  proposed by \citet{efron1992jackknife, efron2012model} that are based on the jackknife and the infinitesimal jackknife (IJ). In practice, bagged predictors are computed using a finite number $B$ of bootstrap replicates, and working with a large $B$ can be computationally expensive. Direct applications of jackknife and IJ estimators to bagging require $B = \ooth(n^{1.5})$ bootstrap replicates to converge, where $n$ is the size of the training set. We propose improved versions that only require $B = \ooth(n)$ replicates. Moreover, we show that the IJ estimator requires 1.7 times less bootstrap replicates than the jackknife to achieve a given accuracy. Finally, we study the sampling distributions of the jackknife and IJ variance estimates themselves. We illustrate our findings with multiple experiments and simulation studies.

\end{abstract}

\section{Introduction}

Bagging \citep{breiman1996bagging} is a popular technique for stabilizing statistical learners. Bagging is often conceptualized as a variance reduction technique, and so it is important to understand how the sampling variance of a bagged learner compares to the variance of the original learner. In this paper, we develop and study methods for estimating the variance of bagged predictors and random forests \citep{breiman2001random}, a popular extension of bagged trees. These variance estimates only require the bootstrap replicates that were used to form the bagged prediction itself, and so can be obtained with moderate computational overhead. The results presented here build on the jackknife-after-bootstrap methodology introduced by \citet{efron1992jackknife} and on the infinitesimal jackknife for bagging (IJ) \citep{efron2012model}. 

\begin{figure}[t]
\centering
\includegraphics[width = 0.6\columnwidth]{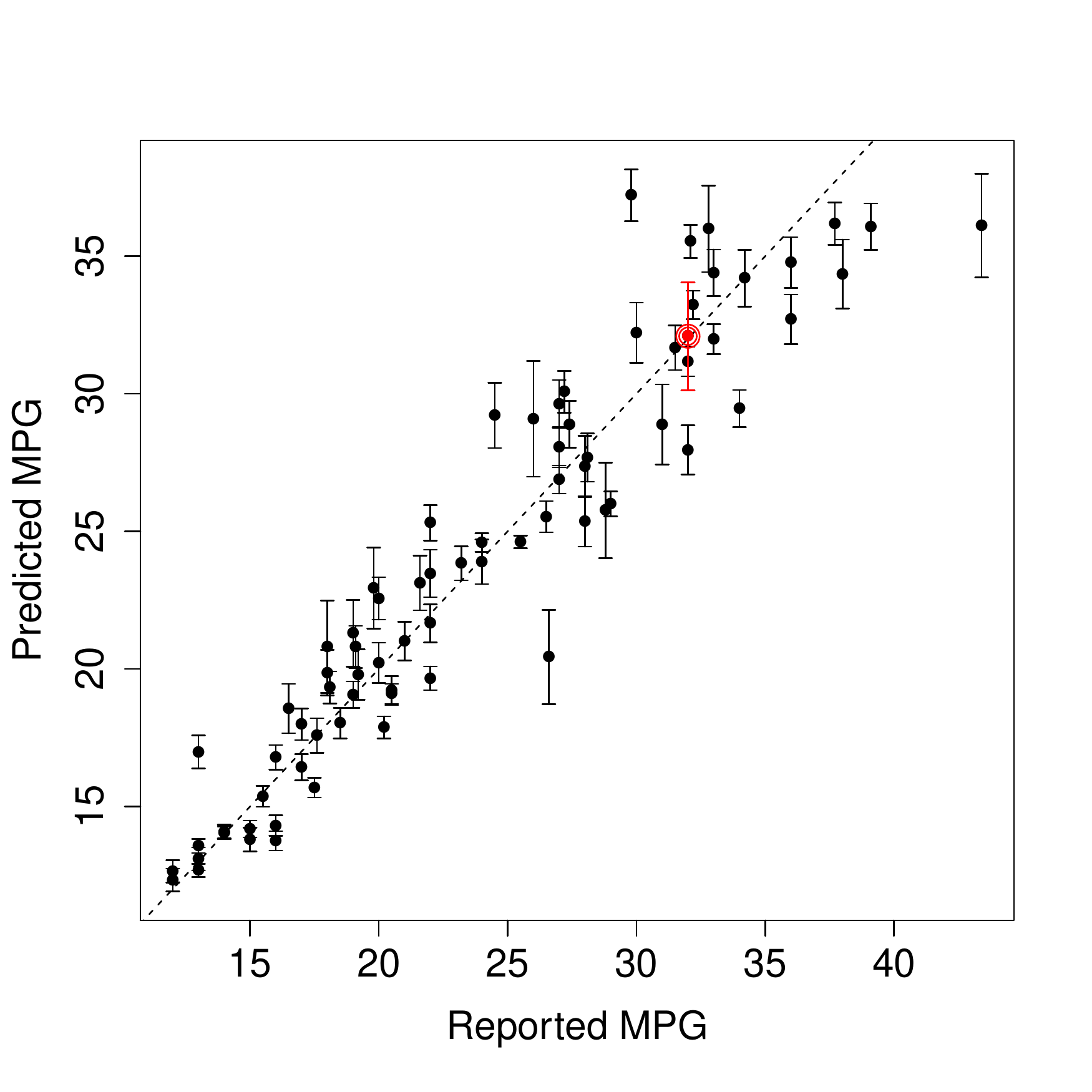}
\caption{Random forest predictions on the ``Auto MPG'' dataset. The random forest was trained using 314 examples; the graph shows results on a test set of size 78. The error bars are 1 standard error in each direction. Because this is a fairly small dataset, we estimated standard errors for the random forest using the averaged estimator from Section \ref{sec:bias}. A more detailed description of the experiment is provided in Appendix \ref{sec:details}.}
\label{fig:auto}
\end{figure}

Figure \ref{fig:auto} shows the results from applying our method to a random forest trained on the ``Auto MPG'' dataset, a regression task where we aim to predict the miles-per-gallon (MPG) gas consumption of an automobile based on 7 features including weight and horsepower.
The error bars shown in Figure \ref{fig:auto} give an estimate of the sampling variance of the random forest; in other words, they tell us how much the random forest's predictions might change if we trained it on a new training set. The fact that the error bars do not in general cross the prediction-equals-observation diagonal suggests that there is some residual noise in the MPG of a car that cannot be explained by a random forest model based on the available predictor variables.\footnote{Our method produces standard error estimates $\hsigma$ for random forest predictions. We then represent these standard error estimates as Gaussian confidence intervals $\hy \pm z_\alpha \hsigma$, where $z_\alpha$ is a quantile of the normal distribution.}

Figure \ref{fig:auto} tells us that the random forest was more confident about some predictions than others.
Rather reassuringly, we observe that the random forest was in general less confident about the predictions for which the reported MPG and predicted MPG were very different. 
There is not a perfect correlation, however, between the error level and the size of the error bars. One of the points, circled in red near (32, 32), appears particularly surprising: the random forest got the prediction almost exactly right, but gave the prediction large errorbars of $\pm 2$. This curious datapoint corresponds to the 1982 Dodge Rampage, a two-door Coupe Utility that is a mix between a passenger car and a truck with a cargo tray. Perhaps our random forest had a hard time confidently estimating the mileage of the Rampage because it could not quite decide whether to cluster it with cars or with trucks. We present experiments on larger datasets in Section \ref{sec:rf}.

Estimating the variance of bagged learners based on the preexisting bootstrap replicates can be challenging, as there are two distinct sources of noise. In addition to the sampling noise (i.e., the noise arising from randomness during data collection), we also need to control the Monte Carlo noise arising from the use of a finite number of bootstrap replicates. We study the effects of both sampling noise and Monte Carlo noise.

In our experience, the errors of the jackknife and IJ estimates of variance are often dominated by Monte Carlo effects. Monte Carlo bias can be particularly troublesome: if we are not careful, the jackknife and IJ estimators can conflate Monte Carlo noise with the underlying sampling noise and badly overestimate the sampling variance. We show how to estimate the magnitude of this Monte Carlo bias and develop bias-corrected versions of the jackknife and IJ estimators that outperform the original ones. We also show that the IJ estimate of variance is able to use the preexisting bootstrap replicates more efficiently than the jackknife estimator by having a lower Monte Carlo variance, and needs 1.7 times less bootstrap replicates than the jackknife to achieve a given accuracy.

If we take the number of bootstrap replicates to infinity, Monte Carlo effects disappear and only sampling errors remain. We compare the sampling biases of both the jackknife and IJ rules and present some evidence that, while the jackknife rule has an upward sampling bias and the IJ estimator can have a downward bias, the arithmetic mean of the two variance estimates can be close to unbiased. We also propose a simple method for estimating the sampling variance of the IJ estimator itself.

Our paper is structured as follows. We first present an overview of our main results in Section \ref{sec:main}, and apply them to random forest examples in Section \ref{sec:rf}.  We then take a closer look at Monte Carlo effects in Section \ref{sec:mc} and analyze the sampling distribution of the limiting IJ and jackknife rules with $B \rightarrow \infty$ in Section \ref{sec:sampling}. We spread simulation experiments throughout Sections \ref{sec:mc} and \ref{sec:sampling} to validate our theoretical analysis.

\subsection{Related Work}

In this paper, we focus on methods based on the jackknife and the infinitesimal jackknife for bagging \citep{efron1992jackknife, efron2012model} that let us estimate standard errors based on the pre-existing bootstrap replicates. Other approaches that rely on forming second-order bootstrap replicates have been studied by \citet{duan2011bootstrap} and \citet{sexton2009standard}. Directly bootstrapping a random forest is usually not a good idea, as it requires forming a large number of base learners. \citet{sexton2009standard}, however, propose a clever work-around to this problem. Their approach, which could have been called a bootstrap of little bags, involves bootstrapping small random forests with around $B = 10$ trees and then applying a bias correction to remove the extra Monte Carlo noise.

There has been considerable interest in studying classes of models for which bagging can achieve meaningful variance reduction, and also in outlining situations where bagging can fail completely \citep[e.g.,][]{skurichina1998bagging, buhlmann2002analyzing, chen2003effects, buja2006observations, friedman2007bagging}. The problem of producing practical estimates of the sampling variance of bagged predictors, however, appears to have received somewhat less attention in the literature so far.

\section{Estimating the Variance of Bagged Predictors}
\label{sec:main}

This section presents our main result: estimates of variance for bagged predictors that can be computed from the same bootstrap replicates that give the predictors. Section \ref{sec:rf} then applies the result to random forests, which can be analyzed as a special class of bagged predictors.

Suppose that we have training examples $Z_1 = (x_1, \, y_1), \, ..., \, Z_n = (x_n, \, y_n)$, an input $x$ to a prediction problem, and a base learner
$\hthRAW(x) = t(x; Z_1, \, ..., \, Z_n).$
To make things concrete, the $Z_i$ could be a list of e-mails $x_i$ paired with labels $y_i$ that catalog the e-mails as either spam or non-spam, $t(x;Z_i)$ could be a decision tree trained on these labeled e-mails, and $x$ could be a new e-mail that we seek to classify. The quantity $\hthRAW(x)$ would then be the output of the tree predictor on input $x$.

With bagging, we aim to stabilize the base learner $t$ by resampling the training data. In our case, the bagged version of $\hthRAW(x)$ is defined as
\begin{equation}
\label{eq:exactbag}
\hthBAGinf(x) = \EE_*[t(x; Z^*_1, \, ..., \, Z^*_n)],
\end{equation}
where the $Z_i^*$ are drawn independently with replacement from the original data (i.e., they form a bootstrap sample). The expectation $\EE_*$ is taken with respect to the bootstrap measure.

The expectation in \eqref{eq:exactbag} cannot in general be evaluated exactly, and so we form the bagged estimator by Monte Carlo
\begin{equation}
\label{eq:bag}
\hthBAG(x) = \frac{1}{B}\sum_{b = 1}^B t^*_b(x), \where t^*_b(x) = t(x; Z^*_{b1}, \, ..., \, Z^*_{bn})
\end{equation} 
and the $Z^*_{bi}$ are elements in the $b^{th}$ bootstrap sample. As $B \rightarrow \infty$, we recover the perfectly bagged estimator $\hthBAGinf(x)$.

\paragraph{Basic variance estimates}
The goal of our paper is to study the sampling variance of bagged learners
$$ V(x) = \pVar{\hthBAGinf(x)}. $$
In other words, we ask how much variance $\hthBAG$ would have once we make $B$ large enough to eliminate the bootstrap effects. We consider two basic estimates of $V$: The \emph{Infinitesimal Jackknife} estimate \citep{efron2012model}, which results in the simple expression
\begin{equation}
\label{eq:IJinf}
\hVIJinf = \sum_{i = 1}^n \Cov_*[N_i^*, \, t^*(x)]^2
\end{equation}
where $\Cov_*[N_i^*, \, t^*(x)]$ is the covariance between $t^*(x)$ and the number of times $N_i^*$ the $i^{th}$ training example appears in a bootstrap sample; and the \emph{Jackknife-after-Bootstrap} estimate \citep{efron1992jackknife}
\begin{equation}
\label{eq:JACKinf}
\hVJinf = \frac{n-1}{n} \sum_{i = 1}^n \left(\bar{t}^*_{(-i)}(x) - \bar{t}^*(x)\right)^2,
\end{equation}
where $\bar{t}^*_{(-i)}(x)$ is the average of $t^*(x)$ over all the bootstrap samples not containing the $i^{th}$ example and $\bar{t}^*(x)$ is the mean of all the $t^*(x)$.

The jackknife-after-bootstrap estimate $\hVJinf$ arises directly by applying the jackknife to the bootstrap distribution. The infinitesimal jackknife \citep{jaeckel1972infinitesimal}, also called the non-parametric delta method, is an alternative to the jackknife where, instead of studying the behavior of a statistic when we remove one observation at a time, we look at what happens to the statistic when we individually down-weight each observation by an infinitesimal amount. When the infinitesimal jackknife is available, it sometimes gives more stable predictions than the regular jackknife. \citet{efron2012model} shows how an application of the infinitesimal jackknife principle to the bootstrap distribution leads to the simple estimate $\hVIJinf$.

\paragraph{Finite-B bias}
In practice, we can only ever work with a finite number $B$ of bootstrap replicates. The natural Monte Carlo approximations to the estimators introduced above are
\begin{equation}
\label{eq:IJ}
 \hVIJ = \sum_{i = 1}^n \widehat{\Cov}_{i}^2 \with \widehat{\Cov}_{i} = \frac{\sum_{b} (N_{bi}^* - 1) (t_b^*(x) - \bar{t}^*(x))}{B},
\end{equation}
and
\begin{align}
\label{eq:JACK}
\hVJ = \frac{n-1}{n} \sum_{i = 1}^n \hDelta_i^2,
&\where \hDelta_i = \hthBAG_{(-i)}(x) - \hthBAG(x) \\
\notag
&\eqand \hthBAG_{(-i)}(x) = \frac{\sum_{\{b \, : \, N^*_{bi} = 0\}} t_b^*(x)}{\left\lvert\{N^*_{bi} = 0\}\right\rvert}.
\end{align}
Here, $N_{bi}^*$ indicates the number of times the $i^{th}$ observation appears in the bootstrap sample $b$.

In our experience, these finite-$B$ estimates of variance are often badly biased upwards if the number of bootstrap samples $B$ is too  small. Fortunately, bias-corrected versions are available:
\begin{align}
\label{eq:IJ-U_teaser}
&\hVIJU = \hVIJ - \frac{n}{B^2}\sum_{b = 1}^B (t_b^*(x) - \bar{t}^*(x))^2, \text{ and } \\
\label{eq:JACK-U_teaser}
&\hVJU = \hVJ - (e - 1) \, \frac{n}{B^2}\sum_{b = 1}^B (t_b^*(x) - \bar{t}^*(x))^2.
\end{align}
These bias corrections are derived in Section \ref{sec:mc}. In many applications, the simple estimators \eqref{eq:IJ} and \eqref{eq:JACK} require $B = \ooth(n^{1.5})$ bootstrap replicates to reduce Monte Carlo noise down to the level of the inherent sampling noise, whereas our bias-corrected versions only require $B = \ooth(n)$ replicates. The bias-corrected jackknife \eqref{eq:JACK-U_teaser} was also discussed by \citet{sexton2009standard}.

\begin{figure}[t]
\centering
\includegraphics[width = 0.5\columnwidth]{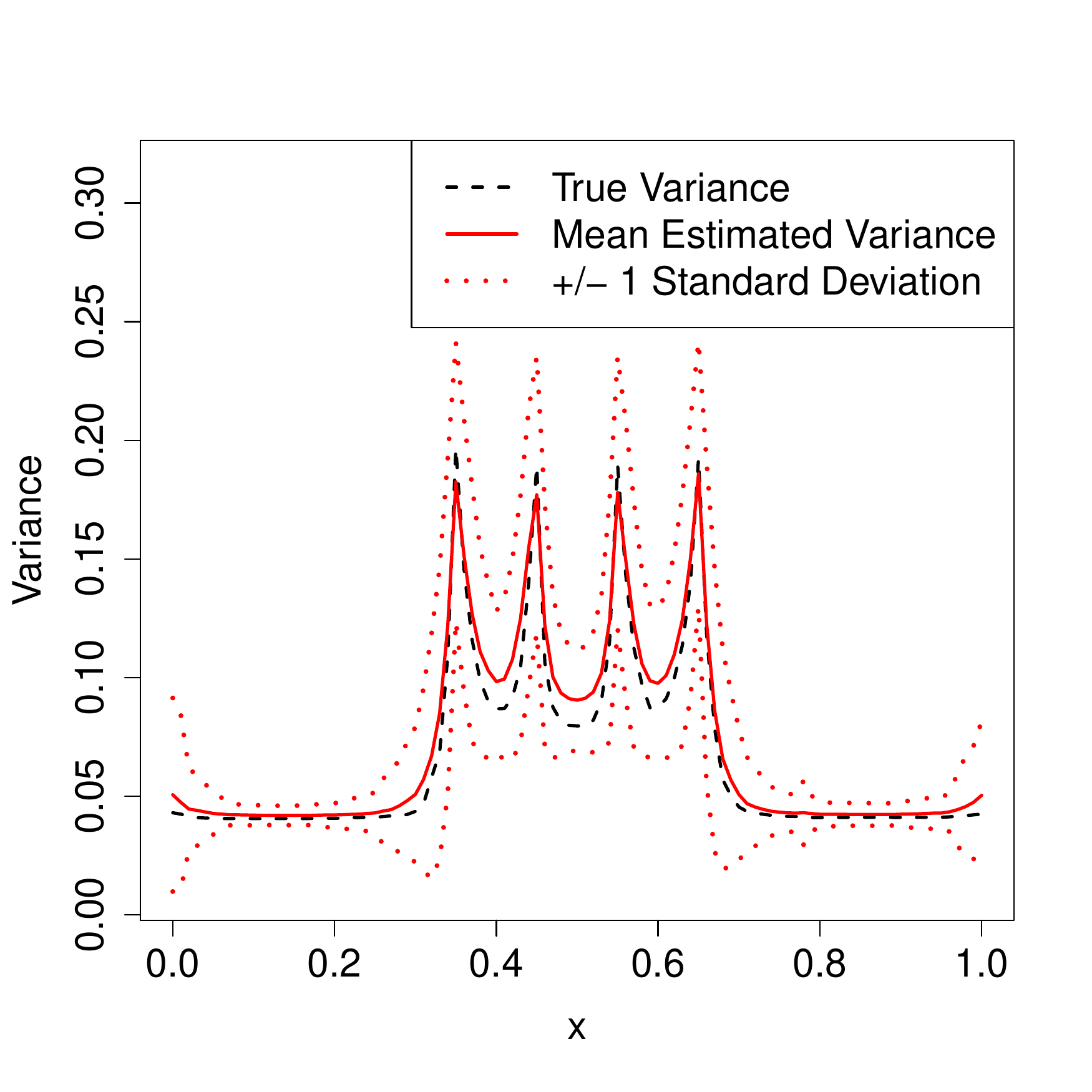}
\caption{Testing the performance of the bias-corrected infinitesimal jackknife estimate of variance for bagged predictors, as defined in \eqref{eq:IJ-U}, on a bagged regression tree. We compare the true sampling error with the average standard error estimate produced by our method across multiple runs; the dotted lines indicate one-standard-error-wide confidence bands for our standard error estimate.}
\label{fig:example}
\end{figure}

In Figure \ref{fig:example}, we show how $\hVIJU$ can be used to accurately estimate the variance of a bagged tree. We compare the true sampling variance of a bagged regression tree with our variance estimate. The underlying signal is a step function with four jumps that are reflected as spikes in the variance of the bagged tree. On average, our variance estimator accurately identifies the location and magnitude of these spikes.

\begin{figure}[t]
\centering
\includegraphics[width = 0.7\columnwidth]{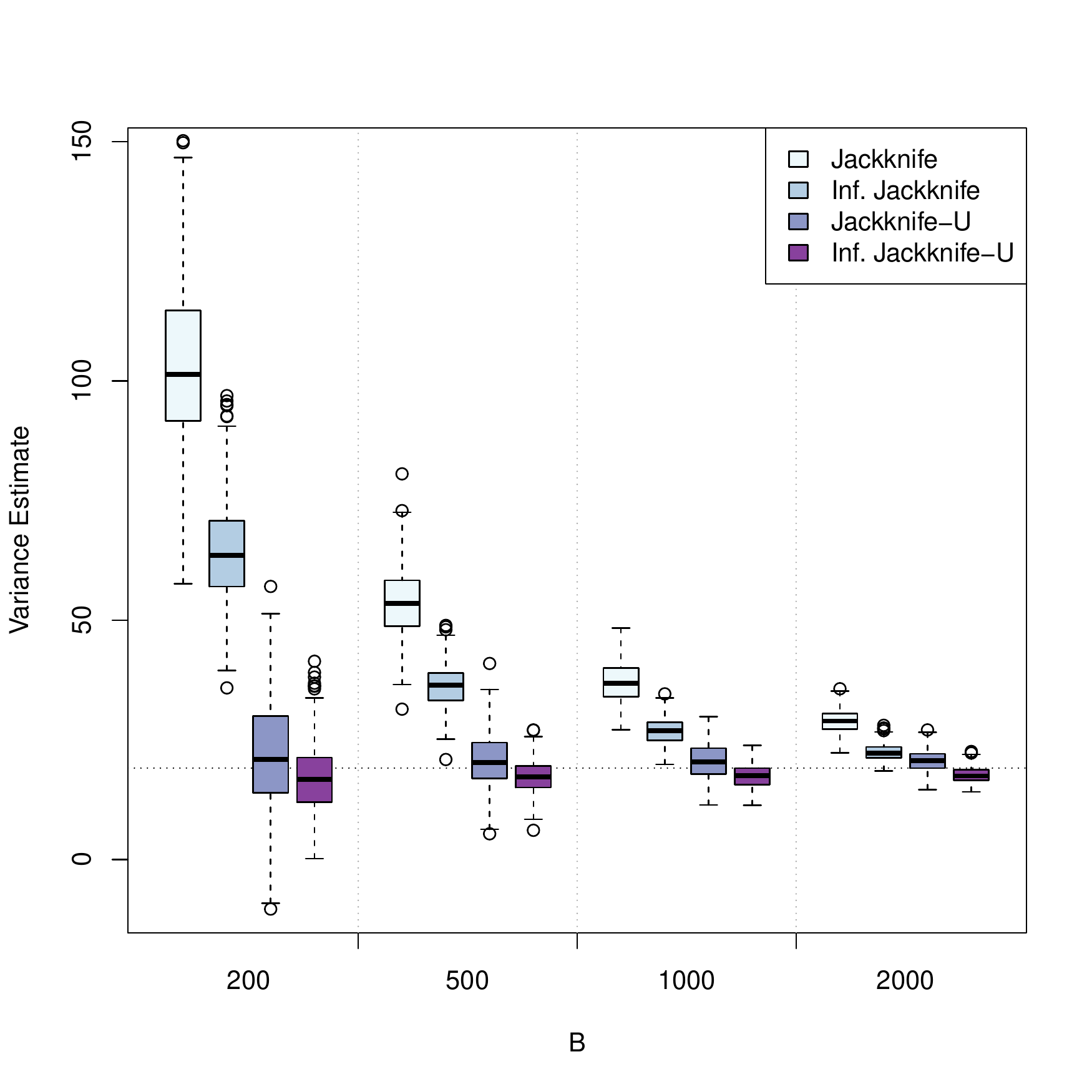}
\caption{Performance, as a function of $B$, of the jackknife and IJ estimators and their bias-corrected modifications on the cholesterol dataset of \citet{efron1991compliance}. The boxplots depict bootstrap realizations of each estimator. The dotted line indicates the mean of all the realizations of the IJ-U and J-U estimators (weighted by $B$).}
\label{fig:chol}
\end{figure}

Figure \ref{fig:chol} compares the performance of the four considered variance estimates on a bagged adaptive polynomial regression example described in detail in Section \ref{sec:mc_example}. We see that the uncorrected estimators $\hVJ$ and $\hVIJ$ are badly biased: the lower whiskers of their boxplots do not even touch the limiting estimate with $B \rightarrow \infty$. We also see that that $\hVIJU$ has about half the variance of $\hVJU$. This example highlights the importance of using estimators that use available bootstrap replicates efficiently: with $B = 500$ bootstrap replicates, $\hVIJU$ can give us a reasonable estimate of $V$, whereas $\hVJ$ is quite unstable and biased upwards by a factor 2.

The figure also suggests that the Monte Carlo noise of $\hVIJ$ decays faster (as a function of $B$) than that of $\hVJ$. This is no accident: as we show in Section \ref{sec:mc_cmp}, the infinitesimal jackknife requires 1.7 times less bootstrap replicates than the jackknife to achieve a given level of level of Monte Carlo error.

\paragraph{Limiting sampling distributions}
The performance of $\hVJ$ and $\hVIJ$ depends on both sampling noise and Monte Carlo noise. In order for $\hVJ$ (and analogously $\hVIJ$) to be accurate, we need both the sampling error of $\hVJinf$, namely $\hVJinf - V$,  and the Monte Carlo error $\hVJ - \hVJinf$ to be small.

It is well known that jackknife estimates of variance are in general biased upwards \citep{efron1981jackknife}. This phenomenon also holds for bagging: $\hVJinf$ is somewhat biased upwards for $V$. We present some evidence suggesting that $\hVIJinf$ is biased downwards by a similar amount, and that the arithmetic mean of $\hVJinf$ and $\hVIJinf$ is closer to being unbiased for $V$ than either of the two estimators alone.

We also develop a simple estimator for the variance of $\hVIJinf$ itself:
\begin{equation*}
\widehat{\pVar{\hVIJinf}} = \sum_{i = 1}^n \left(C^{*, 2}_i - \overline{C^{*, 2}_i}\right)^2,
\end{equation*}
where $C^{*}_i = \Cov_*[N_{bi}^*, \, t_b^*(x)]$ and $\overline{C^{*, 2}_i}$ is the mean of the $C^{*, 2}_i$. 

\section{Random Forest Experiments}
\label{sec:rf}

Random forests \citep{breiman2001random} are a widely used extension of bagged trees. Suppose that we have a tree-structured predictor $t$ and training data $Z_1, \, ..., \, Z_n$. Using notation from \eqref{eq:bag}, the bagged version of this tree predictor is
$$ \hthBAG(x) = \frac{1}{B}\sum_{b = 1}^B t^*_b(x; \, Z_{b1}^*, \, ..., \, Z_{bn}^*). $$
Random forests extend bagged trees by allowing the individual trees $t^*_b$ to depend on an auxiliary noise source $\xi_b$. The main idea is that the auxiliary noise $\xi_b$ encourages more diversity among the individual trees, and allows for more variance reduction than bagging. Several variants of random forests have been analyzed theoretically by, e.g., \citet{biau2008consistency}, \citet{biau2012analysis}, \citet{lin2006random}, and \citet{meinshausen2006quantile}.

Standard implementations of random forests use the auxiliary noise $\xi_b$ to randomly restrict the number of variables on which the bootstrapped trees can split at any given training step. At each step, $m$ features are randomly selected from the pool of all $p$ possible features and the tree predictor must then split on one of these $m$ features. If $m = p$ the tree can always split on any feature and the random forest becomes a bagged tree; if $m = 1$, then the tree has no freedom in choosing which feature to split on.

Following \citet{breiman2001random}, random forests are usually defined more abstractly for theoretical analysis: any predictor of the form
\begin{equation}
\label{eq:rf}
\hthRF(x) = \frac{1}{B} \sum_{b = 1}^B t_b^*(x; \, \xi_b, \, Z_{b1}^*, \, ..., \, Z_{bn}^*) \text{ with } \xi_b \simiid \Xi
\end{equation}
is called a \emph{random forest}. Various choices of noise distribution $\Xi$ lead to different random forest predictors. In particular, trivial noise sources are allowed and so the class of random forests includes bagged trees as a special case. In this paper, we only consider random forests of type \eqref{eq:rf} where individual trees are all trained on bootstrap samples of the training data. We note, however, that that variants of random forests that do not use bootstrap noise have also been found to work well \citep[e.g.,][]{dietterich2000experimental,geurts2006extremely}.

All our results about bagged predictors apply directly to random forests. The reason for this is that random forests can also be defined as bagged predictors with different base learners. Suppose that, on each bootstrap replicate, we drew $K$ times from the auxiliary noise distribution $\Xi$ instead of just once. This would give us a predictor of the form
$$ \hthRF(x) = \frac{1}{B} \sum_{b = 1}^B \frac{1}{K} \sum_{k = 1}^K t_b^*(x; \, \xi_{kb}, \, Z_{b1}^*, \, ..., \, Z_{bn}^*) \text{ with } \xi_{kb} \simiid \Xi. $$
Adding the extra draws from $\Xi$ to the random forest does not change the $B \rightarrow \infty$ limit of the random forest. If we take $K \rightarrow \infty$, we effectively marginalize over the noise from $\Xi$, and get a predictor
\begin{align*}
&\htheta^{\widetilde{RF}}(x) = \frac{1}{B}\sum_{b = 1}^B \tilde{t}^*_b(x; \, Z_{b1}^*, \, ..., \, Z_{bn}^*), \text{ where } \\
&\tilde{t}(x; \, Z_1, \, ..., \, Z_n) = \pEEsub{\xi \sim \Xi}{ t(x; \, \xi, \, Z_1, \, ..., \, Z_n)}.
\end{align*}
In other words, the random forest $\hthRF$ as defined in \eqref{eq:rf} is just a noisy estimate of a bagged predictor with base learner $\tilde{t}$.

It is straight-forward to check that our results about $\hVIJ$ and $\hVJ$ also hold for bagged predictors with randomized base learners. The extra noise from using $t(\cdot; \, \xi)$ instead of $\tilde{t}(\cdot)$ does not affect the limiting correlations in \eqref{eq:IJinf} and \eqref{eq:JACKinf}; meanwhile, the bias corrections from \eqref{eq:IJ-U_teaser} and \eqref{eq:JACK-U_teaser} do not depend on how we produced the $t^*$ and remain valid with random forests. Thus, we can estimate confidence intervals for random forests from $N^*$ and $t^*$ using exactly the same formulas as for bagging.

In the rest of this section, we show how the variance estimates studied in this paper can be used to gain valuable insights in applications of random forests. We use the $\hVIJU$ variance estimate \eqref{eq:IJ-U_teaser} to minimize the required computational resources. We implemented the IJ-U estimator for random forests on top of the \texttt{R} package \texttt{randomForest} \citep{liaw2002classification}.

\subsection{E-mail Spam Example}
\label{sec:email}

The e-mail spam dataset (spambase) is part of a standard classification task, the goal of which is to distinguish spam e-mail (1) from non-spam (0) using $p = 57$ features. Here, we investigate the performance of random forests on this dataset.

\begin{figure}[t]
\centering
\begin{subfigure}{0.325\columnwidth}
\includegraphics[width = \columnwidth]{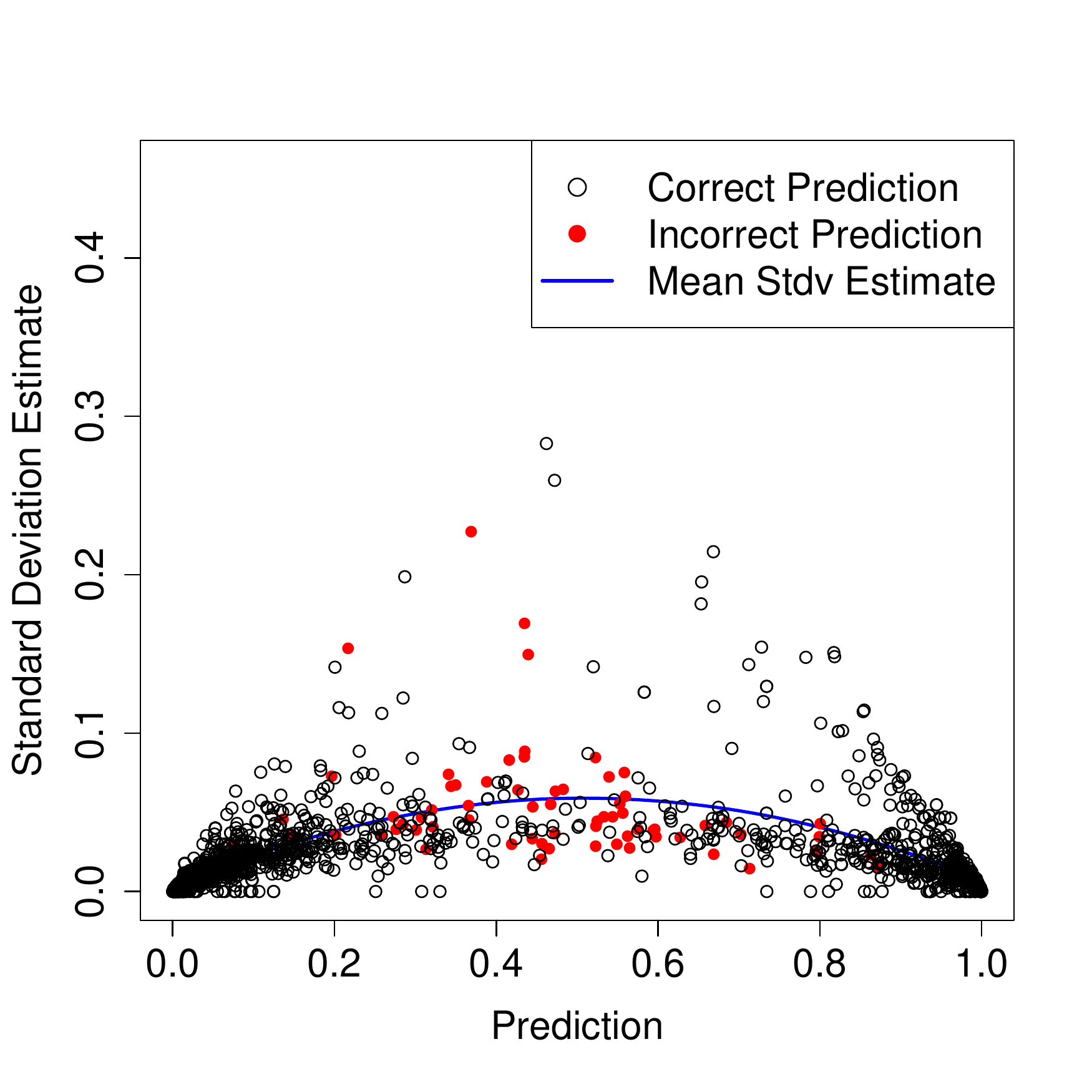}
\caption{$m = 5$}
\end{subfigure}
\begin{subfigure}{0.325\columnwidth}
\includegraphics[width = \columnwidth]{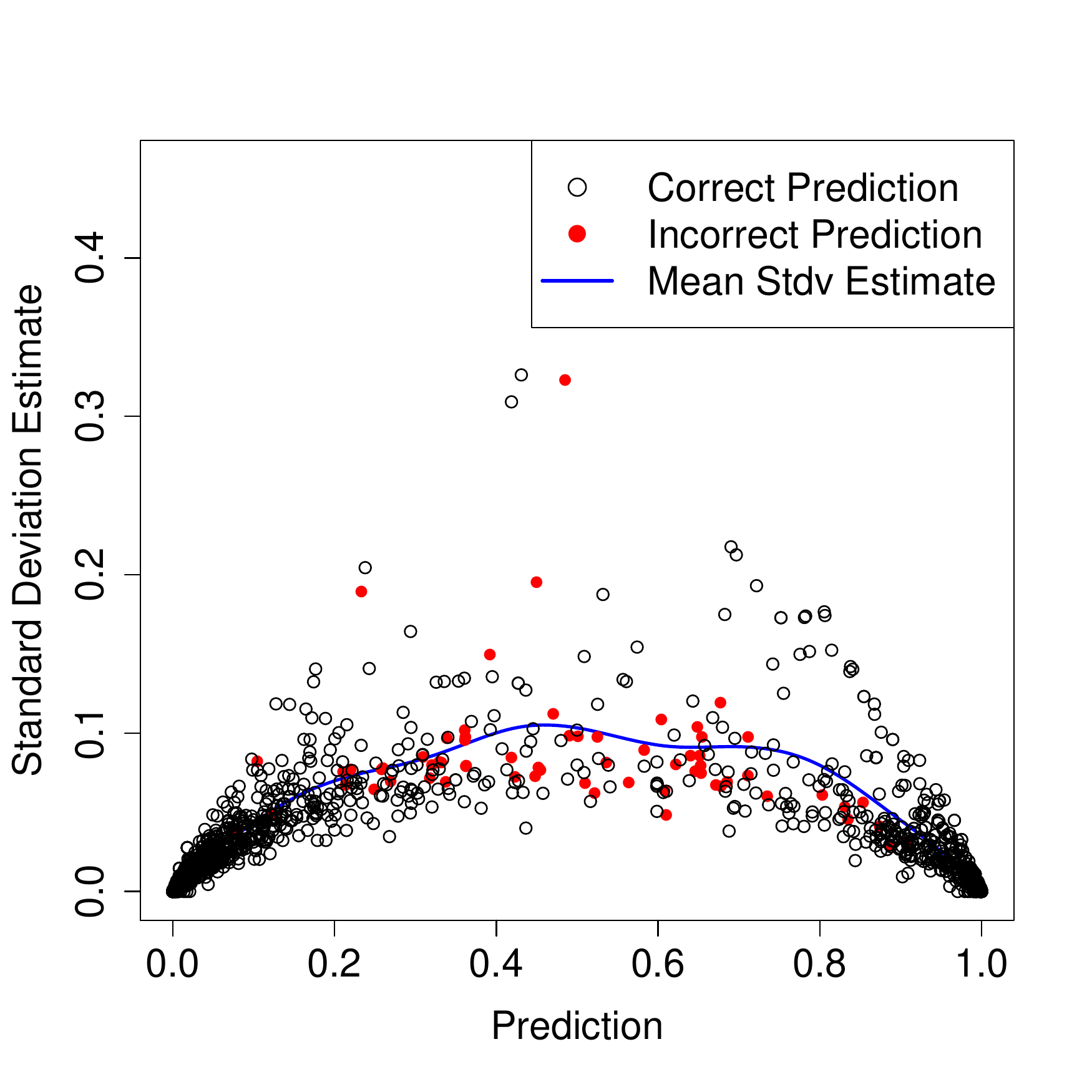}
\caption{$m = 19$}
\end{subfigure}
\begin{subfigure}{0.325\columnwidth}
\includegraphics[width = \columnwidth]{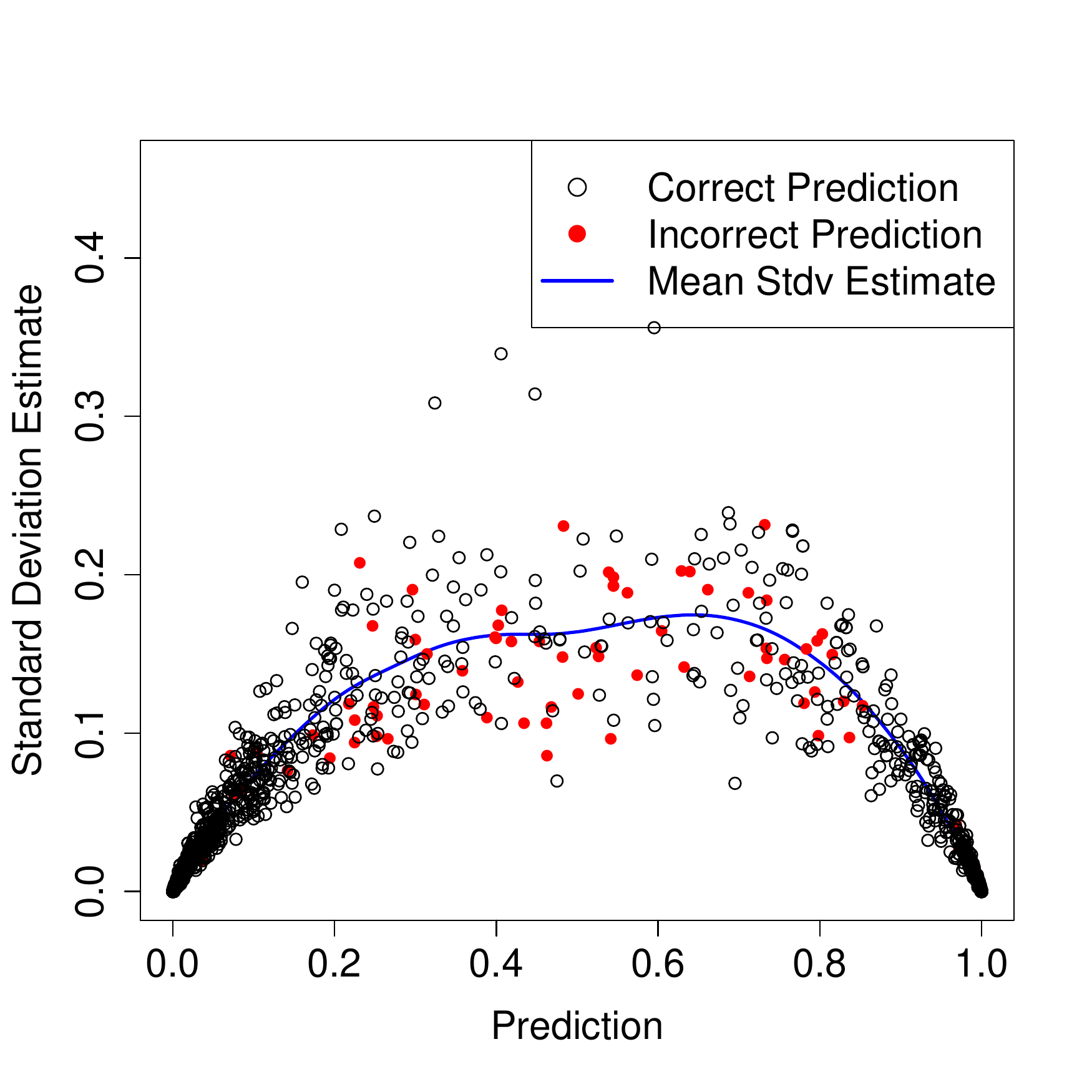}
\caption{$m = 57$ (bagged tree)}
\end{subfigure}
\caption{Standard errors of random forest predictions on the e-mail spam data. The random forests with $m = 5$, $19$, and $57$ splitting variables were all trained on a train set of size $n = 3,065$; the panels above show class predictions and IJ-U estimates for standard errors on a test set of size 1,536. The solid curves are smoothing splines ($df = 4$) fit through the data (including both correct and incorrect predictions).}
\label{fig:spam}
\end{figure}

We fit the spam data using random forests with $m = 5$, $19$ and $57$ splitting variables. With $m = 5$, the trees were highly constrained in their choice of splitting variables, while $m = 57$ is just a bagged tree.
The three random forests obtained test-set accuracies of 95.1\%, 95.2\% and 94.7\% respectively, and it appears that the $m = 5$ or $19$ forests are best. We can use the IJ-U variance formula to gain deeper insight into these numbers, and get a better understanding about what is constraining the accuracy of each predictor.

In Figure \ref{fig:spam}, we plot test-set predictions against IJ-U estimates of standard error for all three random forests. The $m = 57$ random forest appears to be quite unstable, in that the estimated errors are high. Because many of its predictions have large standard errors, it is plausible that the predictions made by the random forest could change drastically if we got more training data. Thus, the $m = 57$ forest appears to suffer from overfitting, and the quality of its predictions could improve substantially with more data.

\begin{figure}[t]
\centering
\includegraphics[width = 0.45\columnwidth]{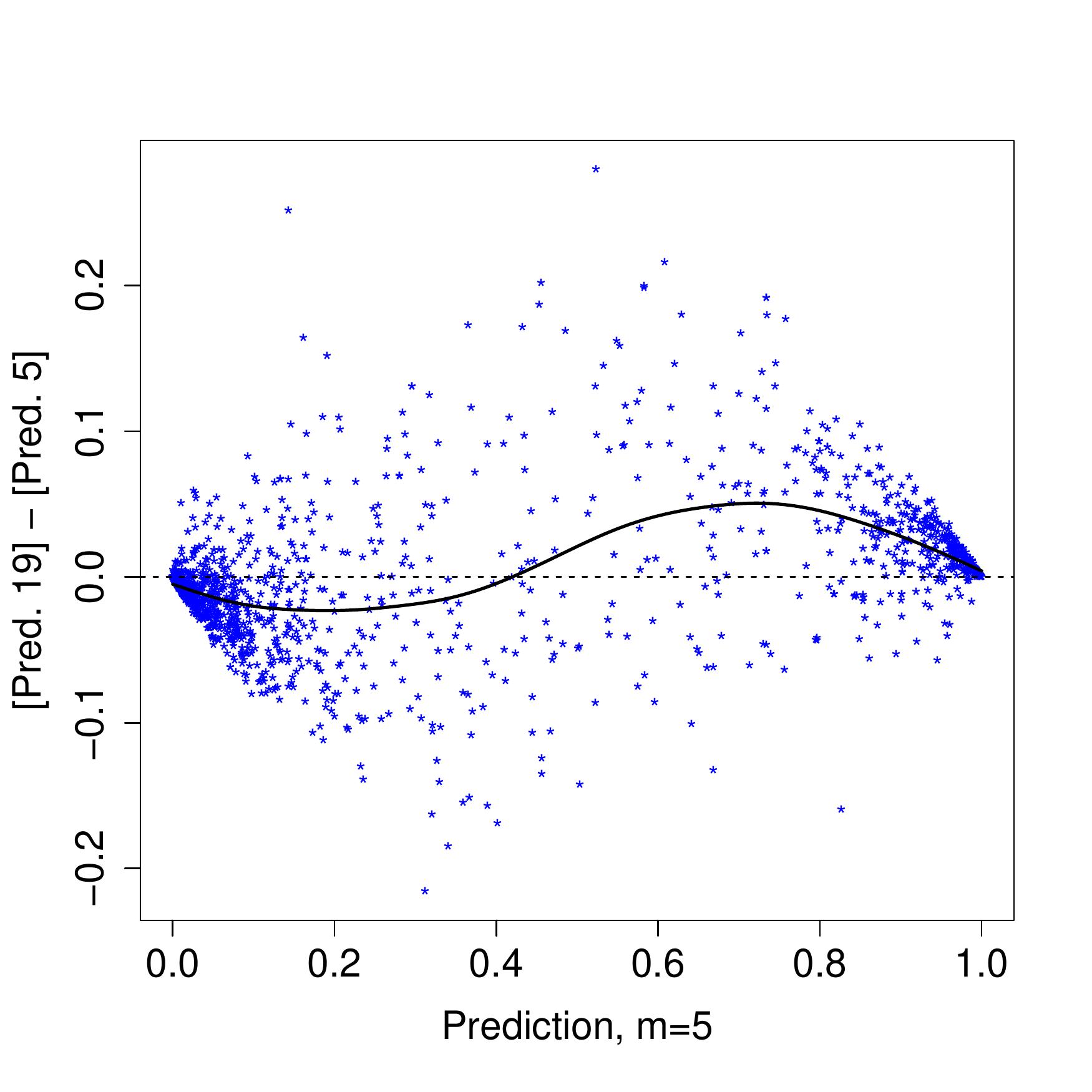}
\caption{Comparison of the predictions made by the $m = 5$ and $m = 19$ random forests. The stars indicate pairs of test set predictions; the solid line is a smoothing spline ($df = 6$) fit through the data.}
\label{fig:spam_cmp}
\end{figure}

Conversely, predictions made by the $m = 5$ random forest appear to be remarkably stable, and almost all predictions have standard errors that lie below 0.1. This suggests that the $m = 5$ forest may be mostly constrained by bias: if the predictor reports that a certain e-mail is spam with probability $0.5 \pm 0.1$, then the predictor has effectively abandoned any hope of unambiguously classifying the e-mail. Even if we managed to acquire much more training data, the class prediction for that e-mail would probably not converge to a strong vote for spam or non-spam.

The $m = 19$ forest appears to have balanced the bias-variance trade-off well. We can further corroborate our intuition about the bias problem faced by the $m = 5$ forest by comparing its predictions with those of the $m = 19$ forest. As shown in Figure \ref{fig:spam_cmp}, whenever the $m = 5$ forest made a cautious prediction that an e-mail might be spam (e.g., a prediction of around 0.8), the $m = 19$ forest made the same classification decision but with more confidence (i.e., with a more extreme class probability estimate $\hat{p}$). Similarly, the $m = 19$ forest tended to lower cautious non-spam predictions made by the $m = 5$ forest. In other words, the $m = 5$ forest appears to have often made lukewarm predictions with mid-range values of $\hat{p}$ on e-mails for which there was sufficient information in the data to make confident predictions. This analysis again suggests that the $m = 5$ forest was constrained by bias and was not able to efficiently use all the information present in the dataset.

\subsection{California Housing Example}
\label{sec:housing}

\begin{figure}[t]
\centering
\begin{subfigure}{0.45\columnwidth}
\includegraphics[width = \columnwidth]{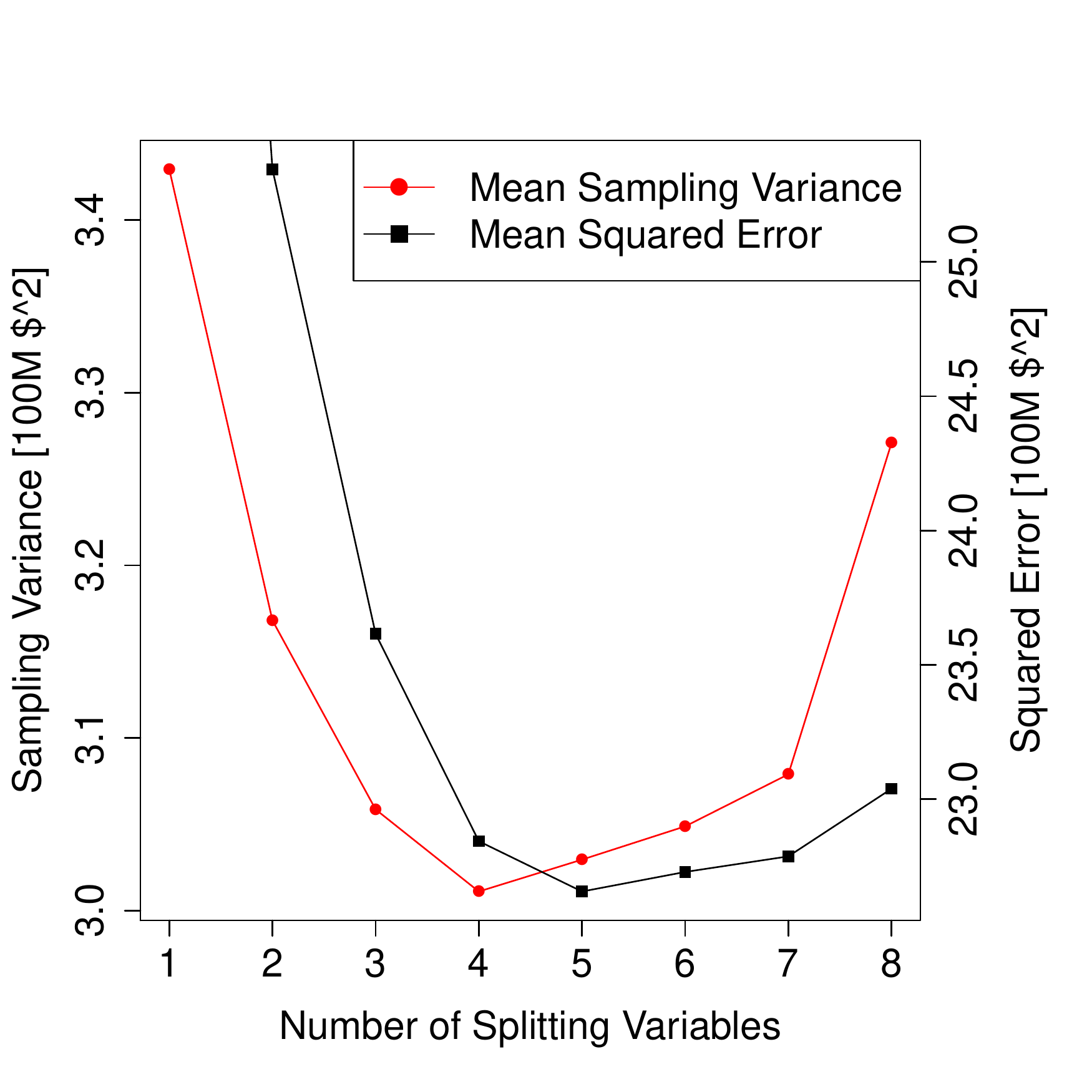}
\caption{MSE and mean variance}
\label{fig:cahousing_mse}
\end{subfigure}
\hspace{0.05\columnwidth}
\begin{subfigure}{0.45\columnwidth}
\includegraphics[width = \columnwidth]{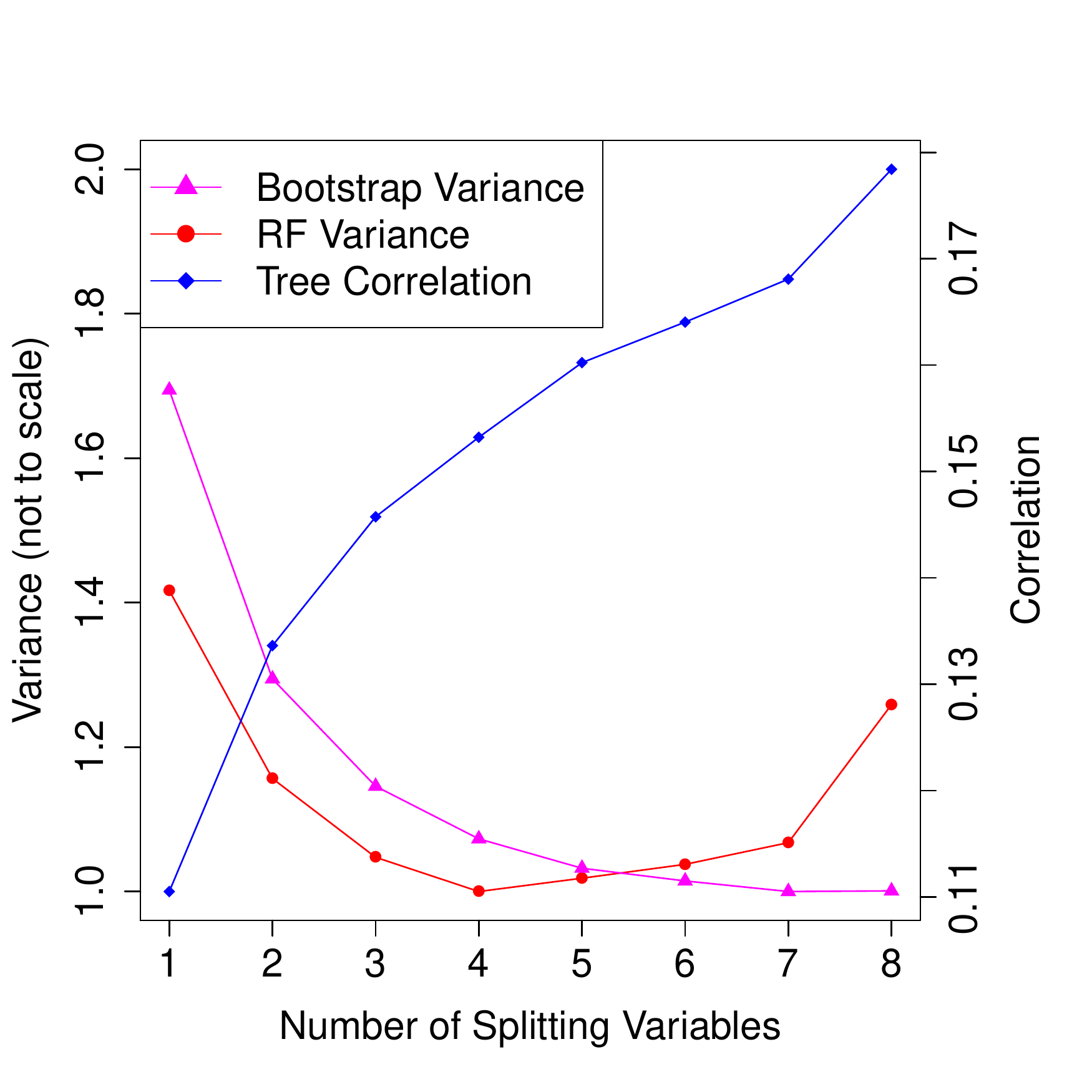}
\caption{Drivers of sampling variance}
\label{fig:cahousing_correl}
\end{subfigure}
\caption{Performance of random forests on the California housing data. The left panel plots MSE and mean sampling variance as a function of the number $m$ of splitting variables. The MSE estimate is the out-of bag error, while the mean sampling variance is the average estimate of variance $\hVIJU$ computed over all training examples. The right panel displays the drivers of sampling variance, namely the variance of the individual bootstrapped trees (bootstrap variance $v$) and their correlation (tree correlation $\rho$).}
\label{fig:housing}
\end{figure}

In the previous example, we saw that the varying accuracy of random forests with different numbers $m$ of splitting variables primarily reflected a bias-variance trade-off. Random forests with small $m$ had high bias, while those with large $m$ had high variance. This bias-variance trade-off does not, however, underlie all random forests. The California housing dataset---a regression task with $n = 20,460$ and $p = 8$---provides a contrasting example.

In Figure \ref{fig:cahousing_mse}, we plot the random forest out-of-bag MSE and IJ-U estimate of average sampling variance across all training examples, with $m$ between 1 and 8. We immediately notice that the sampling variance is not monotone increasing in $m$. Rather, the sampling variance is high if $m$ is too big or too small, and attains a minimum at $m = 4$. Meanwhile, in terms of MSE, the optimal choice is $m = 5$. Thus, there is no bias-variance trade-off here: picking a value of $m$ around 4 or 5 is optimal both from the MSE minimization and the variance minimization points of view.

We can gain more insight into this phenomenon using ideas going back to \citet{breiman2001random}, who showed that the sampling variance of a random forest is governed by two factors: the variance $v$ of the individual bootstrapped trees and their correlation $\rho$. The variance of the ensemble is then $\rho v$. In Figure \ref{fig:cahousing_correl}, we show how both $v$ and $\rho$ react when we vary $m$. Trees with large $m$ are fairly correlated, and so the random forest does not get as substantial a variance reduction over the base learner as with a smaller $m$. With a very small $m$, however, the variance $v$ of the individual trees shoots up, and so the decrease in $\rho$ is no longer sufficient to bring down the variance of the whole forest. The increasing $\rho$-curve and the decreasing $v$-curve thus jointly produce a U-shaped relationship between $m$ and the variance of the random forest. The $m = 4$ forest achieves a low variance by matching fairly stable base learners with a small correlation $\rho$.

\section{Controlling Monte Carlo Error}
\label{sec:mc}

In this section, we analyze the behavior of both the IJ and jackknife estimators under Monte Carlo noise. We begin by discussing the Monte Carlo distribution of the infinitesimal jackknife estimate of variance with a finite $B$; the case of the jackknife-after-bootstrap estimate of variance is similar but more technical and is presented in Appendix \ref{sec:mc_JAB}. We show that the jackknife estimator needs 1.7 times more bootstrap replicates than the IJ estimator to control Monte Carlo noise at a given level. We also highlight a bias problem for both estimators, and recommend a bias correction. When there is no risk of ambiguity, we use the short-hand $t^*$ for $t^*(x)$.

\subsection{Monte Carlo Error for the IJ Estimator}
\label{sec:IJ_mc}

We first consider the Monte Carlo bias of the infinitesimal jackknife for bagging. Let
\begin{equation}
\label{eq:IJdef}
 \hVIJinf = \sum_{i = 1}^n \Cov_*[N_i^*, \, t^*]^2
\end{equation}
be the perfect IJ estimator with $B = \infty$ \citep{efron2012model}. Then, the Monte Carlo bias of $\hVIJ$ is
$$ \pEEsub{*}{\hVIJ} - \hVIJinf = \sum_{i = 1}^n \Var_*[C_i], \where C_i = \frac{\sum_{b} (N_{bi}^* - 1) (t_b^* - \bar{t}^*)}{B} $$
is the Monte Carlo estimate of the bootstrap covariance. Since $t_b^*$ depends on all $n$ observations, $N_{bi}^*$ and $t_b^*$ can in practice be treated as independent for computing $\Var_*[C_i]$, especially when $n$ is large (see remark below).
Thus, as $\Var_*[N_{bi}^*] = 1$, we see that
\begin{equation}
\label{eq:IJ_bias}
\pEEsub{*}{\hVIJ} - \hVIJinf \approx \frac{n\, \vboot}{B}, \where \vboot = \frac{1}{B} \sum_{b = 1}^B (t_b^* - \bar{t}^*)^2.
\end{equation}
Notice that $\vboot$ is the standard bootstrap estimate for the variance of the base learner $\hthRAW(x)$. Thus, the bias of $\hVIJ$ grows linearly in the variance of the original estimator that is being bagged.

Meanwhile, by the central limit theorem, $C_i$ converges to a Gaussian random variable as $B$ gets large. Thus, the Monte Carlo asymptotic variance of $C_i^2$ is approximately $2\Var_*[C_i]^2 + 4\,\EE_*[C_i]^2\Var_*[C_i]$. The $C_i$ can be treated as roughly independent, and so the limiting distribution of the IJ estimate of variance has approximate moments
\begin{equation}
\label{eq:IJ_mom}
\hVIJ - \hVIJinf \approxdot \left(\frac{n\,\vboot}{B}, \, 2\,\frac{n \,\vboot^2}{B^2} + 4\,\frac{\hVIJinf \, \vboot}{B} \right). 
\end{equation}
Interestingly, the Monte Carlo mean squared error (MSE) of $\hVIJ$ mostly depends on the problem through $\vboot$, where $\vboot$ is the bootstrap estimate of the variance of the base learner. In other words, the computational difficulty of obtaining confidence intervals for bagged learners depends on the variance of the base learner.

\subsubsection*{Remark: The IJ Estimator for Sub-bagging}

We have focused on the case where each bootstrap replicate contains exactly $n$ samples. However, in some applications, bagging with subsamples of size $m \neq n$ has been found to work well \citep[e.g.,][]{buhlmann2002analyzing,buja2006observations,friedman2002stochastic,strobl2007bias}. Our results directly extend to the case where $m \neq n$ samples are drawn with replacement from the original sample. We can check that \eqref{eq:IJdef} still holds, but now $\Var{N_{bi}^*} = m/n$. Carrying out the same analysis as above, we can establish an analogue to \eqref{eq:IJ_mom}:
\begin{equation}
\label{eq:IJ_momsub}
\hVIJ(m) - \hVIJinf(m) \approxdot \left(\frac{m\,\vboot}{B}, \, 2\,\frac{m^2 \,\vboot^2}{n B^2} + 4\,\frac{m \hVIJinf \, \vboot}{n B} \right). 
\end{equation}
For simplicity of exposition, we will restrict our analysis to the case $m = n$ for the rest of this paper.

\subsubsection*{Remark: Approximate Independence}
In the above derivation, we used the approximation
$$ \Var_*\left[(N_{bi}^* - 1) (t_b^* - \bar{t}^*)\right] \approx \Var_*\left[N_{bi}^*\right]\Var_*\left[t_b^*\right]. $$
We can evaluate the accuracy of this approximation using the formula
\begin{align*}
&\Var_*\left[(N_{bi}^* - 1) (t_b^* - \bar{t}^*)\right] - \Var_*\left[N_{bi}^*\right]\Var\left[t_b^*\right] \\
&\ \ \ \ = \Cov_*\left[(N_{bi}^* - 1)^2, \, (t_b^* - \bar{t}^*)^2\right] - \Cov_*\left[(N_{bi}^* - 1), \, (t_b^* - \bar{t}^*)\right]^2.
\end{align*}
In the case of the sample mean $t(Z_1^*, \, ..., \, Z_n^*) = \frac{1}{n}\sum_{i} Z_i^*$ paired with the Poisson bootstrap, this term reduces to
$$ \Cov_*\left[(N_{bi}^* - 1)^2, \, (t_b^* - \bar{t}^*)^2\right] - \Cov_*\left[(N_{bi}^* - 1), \, (t_b^* - \bar{t}^*)\right]^2 = 2\frac{\left(Z_i - \bar{Z}\right)^2}{n^2}, $$
and the correction to \eqref{eq:IJ_bias} would be $2\hv / (nB) \ll n\hv/B$.

\subsection{Comparison of Monte Carlo Errors}
\label{sec:mc_cmp}

As shown in Appendix \ref{sec:mc_JAB}, the Monte Carlo error for the jackknife-after-bootstrap estimate of variance has approximate moments
\begin{equation}
\label{eq:JACK_mom}
\hVJ - \hVJinf \approxdot \left( (e -1)\, \frac{n\,\vboot}{B}
, \, 2\, (e-1)^2 \frac{n\, \hv^2}{B^2} + 4  \, (e-1) \, \frac{\hVJinf \, \vboot}{B} \right)
\end{equation}
where $\hVJinf$ is the jackknife estimate computed with $B = \infty$ bootstrap replicates. The Monte Carlo stability of $\hVJ$ again primarily depends on $\vboot$.

By comparing \eqref{eq:IJ_mom} with \eqref{eq:JACK_mom}, we notice that the IJ estimator makes better use of a finite number $B$ of bootstrap replicates than the jackknife estimator. For a fixed value of $B$, the Monte Carlo bias of $\hVJ$ is about $e-1$ or 1.7 times as large as that of $\hVIJ$; the ratio of Monte Carlo variance starts off at 3 for small values of $B$ and decays down to 1.7 as $B$ gets much larger than $n$. Alternatively, we see that the IJ estimate with $B$ bootstrap replicates has errors on the same scale as the jackknife estimate with $1.7 \cdot B$ replicates.

This suggests that if computational considerations matter and there is a desire to perform as few bootstrap replicates $B$ as possible while controlling Monte Carlo error, the infinitesimal jackknife method may be preferable to the jackknife-after-bootstrap.

\subsection{Correcting for Monte Carlo Bias}
\label{sec:mc_bias}

The Monte Carlo MSEs of $\hVIJ$ and $\hVJ$ are in practice dominated by bias, especially for large $n$. Typically, we would like to pick $B$ large enough to keep the Monte Carlo MSE on the order of $1/n$. For both \eqref{eq:IJ_mom} and \eqref{eq:JACK_mom}, we see that performing $B = \ooth(n)$ bootstrap iterations is enough to control the variance. To reduce the bias to the desired level, namely $\oo(n^{-0.5})$, we would need to take $B = \ooth(n^{1.5})$ bootstrap samples.

Although the Monte Carlo bias for both $\hVIJ$ and $\hVJ$ is large, this bias only depends on $\vboot$ and so is highly predictable. This suggests a bias-corrected modification of the IJ and jackknife estimators respectively:
\begin{align}
\label{eq:IJ-U}
&\hVIJU = \hVIJ - \frac{n\,\vboot}{B}, \text{ and } \\
\label{eq:JACK-U}
&\hVJU = \hVJ -  (e - 1) \, \frac{n\,\vboot}{B}.
\end{align}
Here $\hVIJ$ and $\hVJ$ are as defined in \eqref{eq:IJ}, and $\vboot$ is the bootstrap estimate of variance from \eqref{eq:IJ_bias}. The letter U stands for unbiased. This transformation effectively removes the Monte Carlo bias in our experiments without noticeably increasing variance. The bias corrected estimates only need $B = \ooth(n)$ bootstrap replicates to control Monte Carlo MSE at level $1/n$.

\subsection{A Numerical Example}
\label{sec:mc_example}

To validate the observations made in this section, we re-visit the cholesterol dataset used by \citet{efron2012model} as a central example in developing the IJ estimate of variance. The dataset \citep[introduced by][]{efron1991compliance} contains records for $n = 164$ participants in a clinical study, all of whom received a proposed cholesterol-lowering drug. The data contains a measure $d$ of the cholesterol level decrease observed for each subject, as well as a measure $c$ of compliance (i.e. how faithful the subject was in taking the medication). \citeauthor{efron1991compliance} originally fit $d$ as a polynomial function of $c$; the degree of the polynomial was adaptively selected by minimizing \citeauthor{mallows1973some}' $C_p$ criterion \citeyearpar{mallows1973some}.

We here follow \citet{efron2012model} and study the bagged adaptive polynomial fit of $d$ against $c$ for predicting the cholesterol decrease of a new subject with a specific compliance level. The degree of the polynomial is selected among integers between 1 and 6 by $C_p$ minimization. \citet{efron2012model} gives a more detailed description of the experiment. We restrict our attention to predicting the cholesterol decrease of a new patient with compliance level $c = -2.25$; this corresponds to the patient with the lowest observed compliance level.

\begin{figure}[t]
\centering
\includegraphics[width = 0.45\columnwidth]{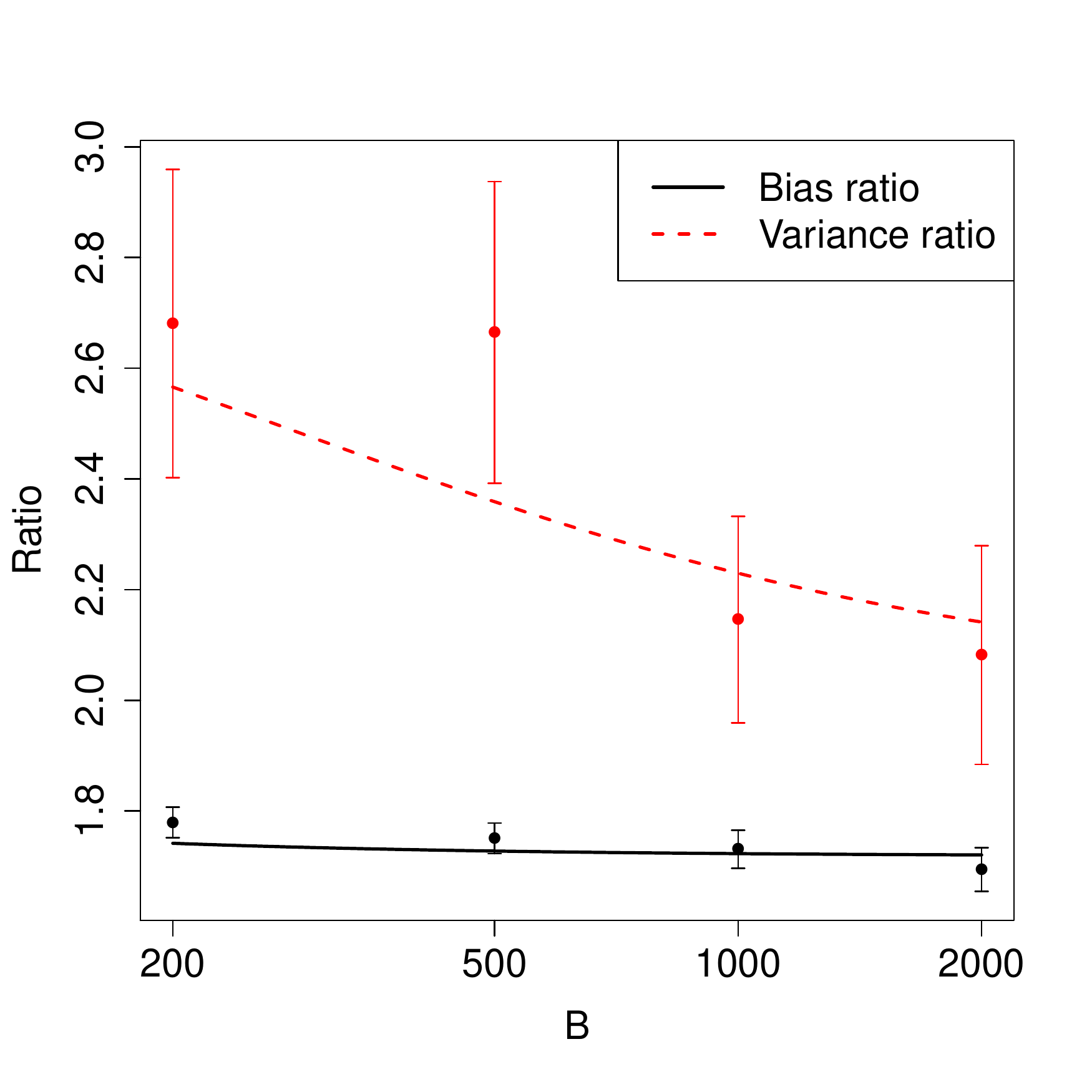}
\caption{Predicted and actual performance ratios for the uncorrected $\hVJ$ and $\hVIJ$ estimators in the cholesterol compliance example. The plot shows both $\Var[\hVJ]/\Var[\hVIJ]$ and $\Bias[\hVJ]/\Bias[\hVIJ]$. The observations are derived from the data presented in Figure \ref{fig:chol}; the error bars are one standard deviation in each direction. The solid lines are theoretical predictions obtained from \eqref{eq:IJ_mom} and \eqref{eq:JACK_mom}.}
\label{fig:chol_ratio}
\end{figure}

In Figure \ref{fig:chol}, we compare the performance of the variance estimates for bagged predictors studied in this paper. The boxplots depict repeated realizations of the variance estimates with a finite $B$. We can immediately verify the qualitative insights presented in this section. Both the jackknife and IJ rules are badly biased for small $B$, and this bias goes away more slowly than the Monte Carlo variance. Moreover, at any given $B$, the jackknife estimator is noticeably less stable than the IJ estimator.

The J-U and IJ-U estimators appear to fix the bias problem without introducing instability. The J-U estimator has a slightly higher mean than the IJ-U one. As discussed in Section \ref{sec:bias}, this is not surprising, as the limiting ($B \rightarrow \infty$) jackknife estimator has an upward sampling bias while the limiting IJ estimator can have a downward sampling bias. The fact that the J-U and IJ-U estimators are so close suggests that both methods work well for this problem.

The insights developed here also appear to hold quantitatively. In Figure \ref{fig:chol_ratio}, we compare the ratios of Monte Carlo bias and variance for the jackknife and IJ estimators with theoretical approximations implied by \eqref{eq:IJ_mom} and \eqref{eq:JACK_mom}. The theoretical formulas appear to present a credible picture of the relative merits of the jackknife and IJ rules.

\section{Sampling Distribution of Variance Estimates}
\label{sec:sampling}

In practice, the $\hVIJ$ and $\hVJ$ estimates are computed with a finite number $B$ of bootstrap replicates. In this section, however, we let $B$ go to infinity, and study the sampling properties of the IJ and jackknife variance estimates in the absence of Monte Carlo errors. In other words, we study the impact of noise in the data itself. Recall that we write $\hVIJinf$ and $\hVJinf$ for the limiting estimators with $B = \infty$ bootstrap replicates.

We begin by developing a simple formula for the sampling variance of $\hVIJinf$ itself. In the process of developing this variance formula, we obtain an ANOVA expansion of $\hVIJinf$ that we then use in Section \ref{sec:bias} to compare the sampling biases of the jackknife and infinitesimal jackknife estimators.

\subsection{Sampling Variance of the IJ Estimate of Variance}
\label{sec:varvar}

If the data $Z_i$ are independently drawn from a distribution $F$, then the variance of the IJ estimator is very nearly given by
\begin{align}
\label{eq:IJ_varvar}
\pVarsub{F}{\hVIJinf} &\approx n\pVarsub{F}{h_F^2(Z)}, \where \\
\label{eq:firstorder}
&h_F(Z) = \pEEsub{F}{\hthBAGinf|Z_1 = Z} - \pEEsub{F}{\hthBAGinf}.
\end{align}
This expression suggests a natural plug-in estimator
\begin{equation}
\label{eq:IJ_varvarhat}
\widehat{\Var[\hVIJ]} = \sum_{i = 1}^n \left(C^{*, 2}_i - \overline{C^{*, 2}_i}\right)^2,
\end{equation}
where $C^{*}_i = \Cov_*[N_{bi}^*, \, t_b^*]$ is a bootstrap estimate for $h_F(Z_i)$ and $\overline{C^{*, 2}_i}$ is the mean of the $C^{*, 2}_i$. The rest of the notation is as in Section \ref{sec:main}.

The relation \eqref{eq:IJ_varvar} arises from a general connection between the infinitesimal jackknife and the theory of H{\'a}jek projections. The H{\'a}jek projection of an estimator is the best approximation to that estimator that only considers first-order effects. In our case, the H{\'a}jek projection of $\hthBAGinf$ is
\begin{equation}
\label{eq:hajek}
\hthBAGinf_H = \pEEsub{F}{\hthBAGinf} + \sum_{i = 1}^n h_F(Z_i),
\end{equation}
where $h_F(Z_i)$ is as in \eqref{eq:firstorder}. The variance of the H\'ajek projection is $\pVar{\hthBAGinf_H} = n\pVar{h_F(Z)}$.

The key insight behind \eqref{eq:IJ_varvar} is that the IJ estimator is effectively trying to estimate the variance of the H\'ajek projection of $\hthBAG$, and that
\begin{equation}
\label{eq:approxhajek}
\hVIJinf \approx \sum_{i = 1}^n h^2_F(Z_i). 
\end{equation}
The approximation \eqref{eq:IJ_varvar} then follows immediately, as the right-hand side of the above expression is a sum of independent random variables. Note that we cannot apply this right-hand side expression directly, as $h$ depends on the unknown underlying distribution $F$.

The connections between H\'ajek projections and the infinitesimal jackknife have been understood for a long time. \citet{jaeckel1972infinitesimal} originally introduced the infinitesimal jackknife as a practical approximation to the first-order variance of an estimator (in our case, the right-hand side of \eqref{eq:approxhajek}). More recently, \citet{efron2012model} showed that $\hVIJinf$ is equal to the variance of a ``bootstrap H\'ajek projection.'' In Appendix \ref{sec:varvartheory}, we build on these ideas and show that, in cases where a plug-in approximation is valid, \eqref{eq:approxhajek} holds very nearly for bagged estimators.

\begin{figure}[t]
\centering
\begin{subfigure}{0.45\columnwidth}
\includegraphics[width = \columnwidth]{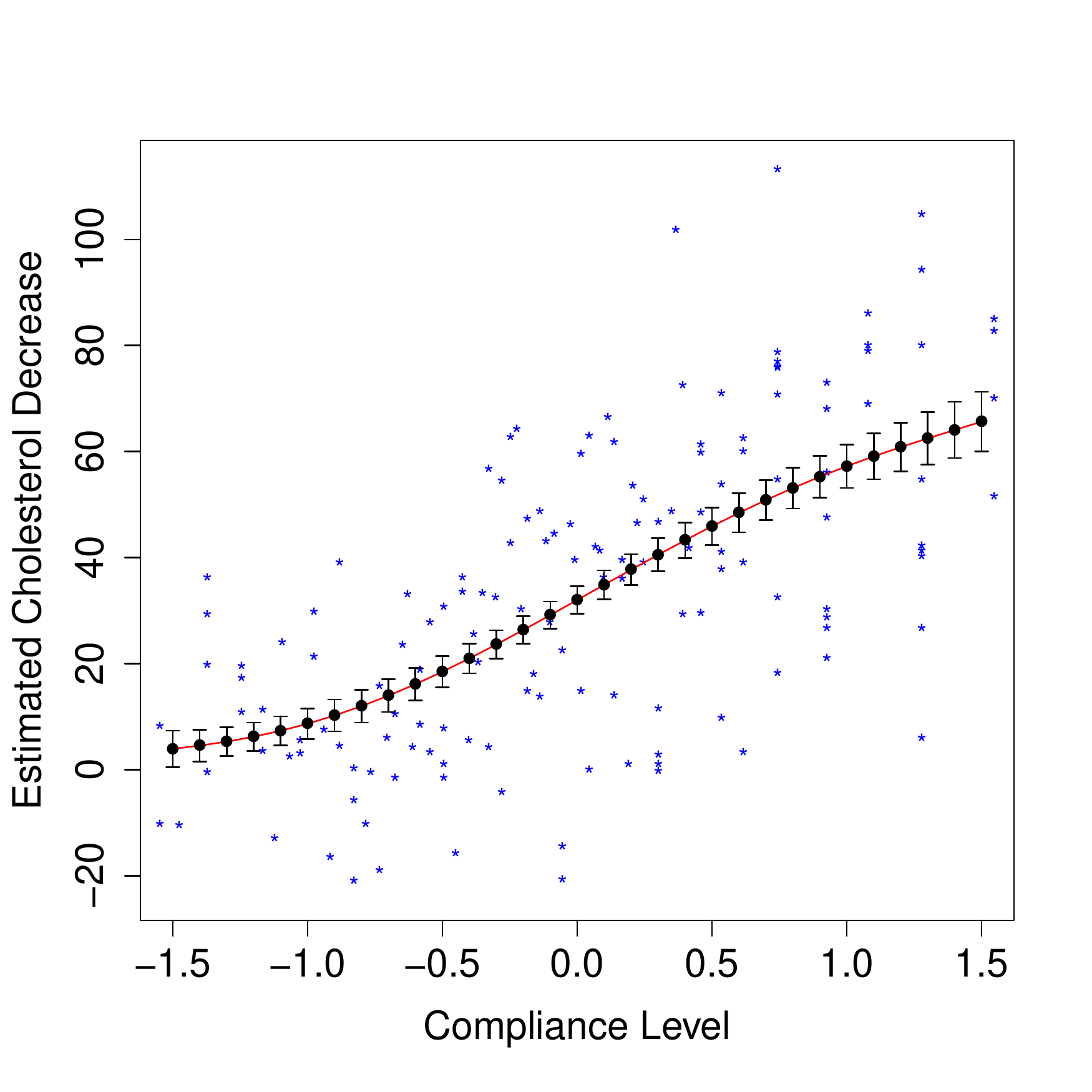}
\caption{Bagged polynomial fit.}
\end{subfigure}
\hspace{0.05\columnwidth}
\begin{subfigure}{0.45\columnwidth}
\includegraphics[width = \columnwidth]{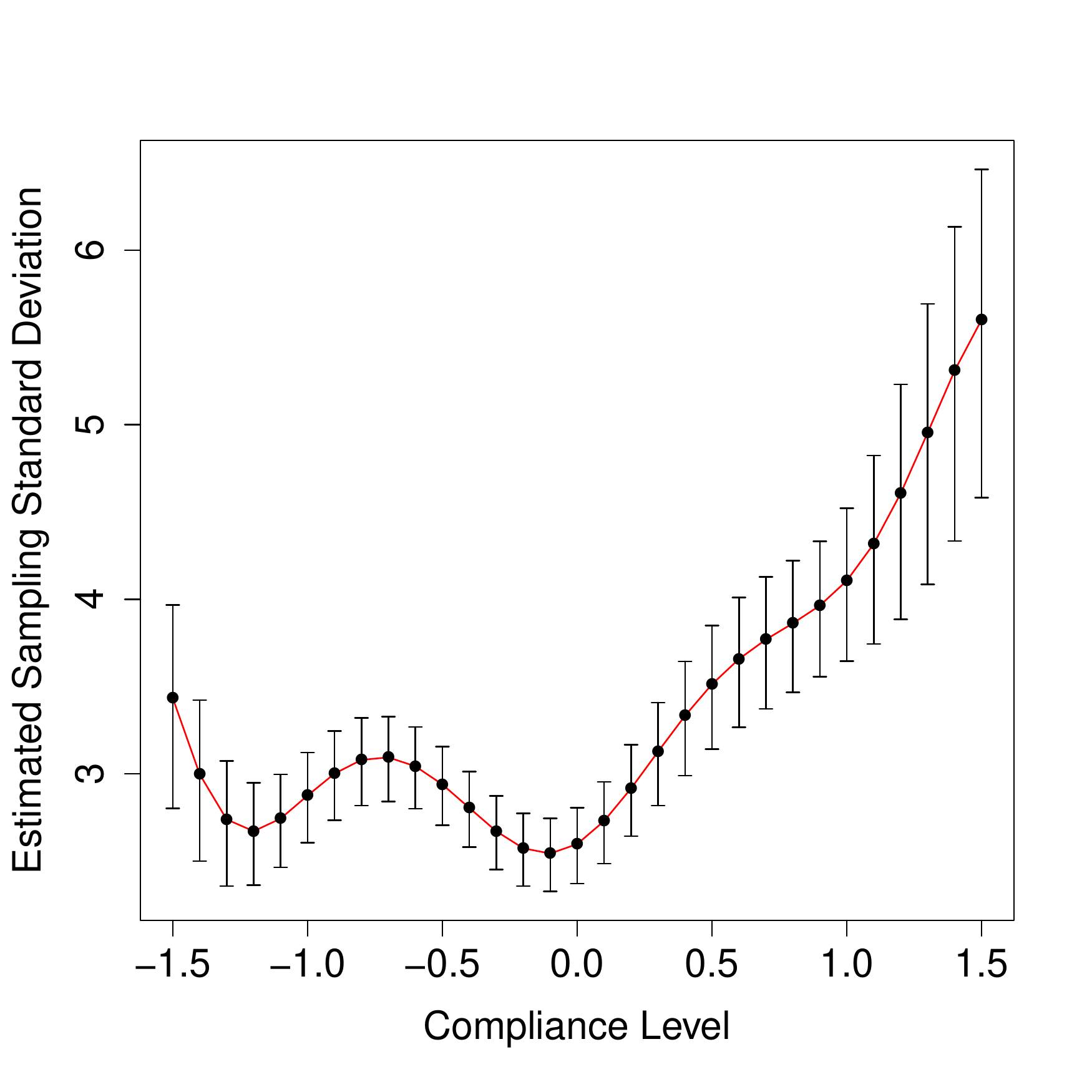}
\caption{Errors for the IJ estimator}
\label{fig:varvarb}
\end{subfigure}
\caption{Stability of the IJ estimate of variance on the cholesterol data. The left panel shows the bagged fit to the data, along with error bars generated by the IJ method; the stars denote the data (some data points have $x$-values that exceed the range of the plot). In the right panel, we use \eqref{eq:IJ_varvarhat} to estimate error bars for the error bars in the first panel. All error bars are one standard deviation in each direction.}
\label{fig:varvar}
\end{figure}

We apply our variance formula to the cholesterol dataset of \citet{efron2012model}, following the methodology described in Section \ref{sec:mc_example}. 
In Figure \ref{fig:varvar}, we use the formula \eqref{eq:IJ_varvarhat} to study the sampling variance of $\hVIJinf$ as a function of the compliance level $c$. The main message here is rather reassuring: as seen in Figure \ref{fig:varvarb}, the coefficient of variation of $\hVIJinf$ appears to be fairly low, suggesting that the IJ variance estimates can be trusted in this example. Note that, the formula from \eqref{eq:IJ_varvarhat} can require many bootstrap replicates to stabilize and suffers from an upward Monte Carlo bias just like $\hVIJ$. We used $B = 100,000$ bootstrap replicates to generate Figure \ref{fig:varvar}.

\subsection{Sampling Bias of the Jackknife and IJ Estimators}
\label{sec:bias}

We can understand the sampling biases of both the jackknife and IJ estimators in the context of the ANOVA decomposition of \citet{efron1981jackknife}. Suppose that we have data $Z_1, \, ..., \, Z_n$ drawn independently from a distribution $F$, and compute our estimate $\hthBAGinf$ based on this data. Then, we can decompose its variance as
\begin{equation}
\label{eq:anova}
\pVarsub{F}{\hthBAGinf} = V_1 + V_2 + ... + V_n,
\end{equation}
where
$$ V_1 = n\pVarsub{F}{\pEEsub{F}{\hthBAGinf | Z_1}} $$
is the variance due to first-order effects, $V_2$ is the variance due to second-order effects of the form
$$\pEEsub{F}{\hthBAGinf | Z_1, \, Z_2} - \pEEsub{F}{\hthBAGinf | Z_1} - \pEEsub{F}{\hthBAGinf | Z_2} + \pEEsub{F}{\hthBAGinf}, $$
and so on. Note that all the terms $V_k$ are non-negative.

\citet{efron1981jackknife} showed that, under general conditions, the jackknife estimate of variance is biased upwards. In our case, their result implies that the jackknife estimator computed on $n+1$ data points has variance
\begin{equation}
\label{eq:JACK_BIAS}
\pEEsub{F}{\hVJinf} = V_1 + 2V_2 + 3V_3 + ... + nV_n.
\end{equation}
Meanwhile, \eqref{eq:approxhajek} suggests that
\begin{equation}
\label{eq:IJ_BIAS}
\pEEsub{F}{\hVIJinf} \approx V_1.
\end{equation}
In other words, on average, both the jackknife and IJ estimators get the first-order variance term right. The jackknife estimator then proceeds to double the second-order term, triple the third-order term etc, while the IJ estimator just drops the higher order terms.

By comparing \eqref{eq:JACK_BIAS} and \eqref{eq:IJ_BIAS}, we see that the upward bias of $\hVJinf$ and the downward bias of $\hVIJinf$ partially cancel out. In fact,
\begin{equation}
\label{eq:MEAN_BIAS}
\pEEsub{F}{\frac{\hVJinf + \hVIJinf}{2}} \approx V_1 + V_2 + \frac{3}{2} V_3 +... + \frac{n}{2}V_n,
\end{equation}
and so the arithmetic mean of $\hVJinf$ and $\hVIJinf$ has an upward bias that depends only on third- and higher-order effects. Thus, we might expect that in small-sample situations where $\hVJinf$ and $\hVIJinf$ exhibit some bias, the mean of the two estimates may work better than either of them taken individually.

\begin{figure}[t]
\centering
\begin{subfigure}{0.45\columnwidth}
\includegraphics[width = \columnwidth]{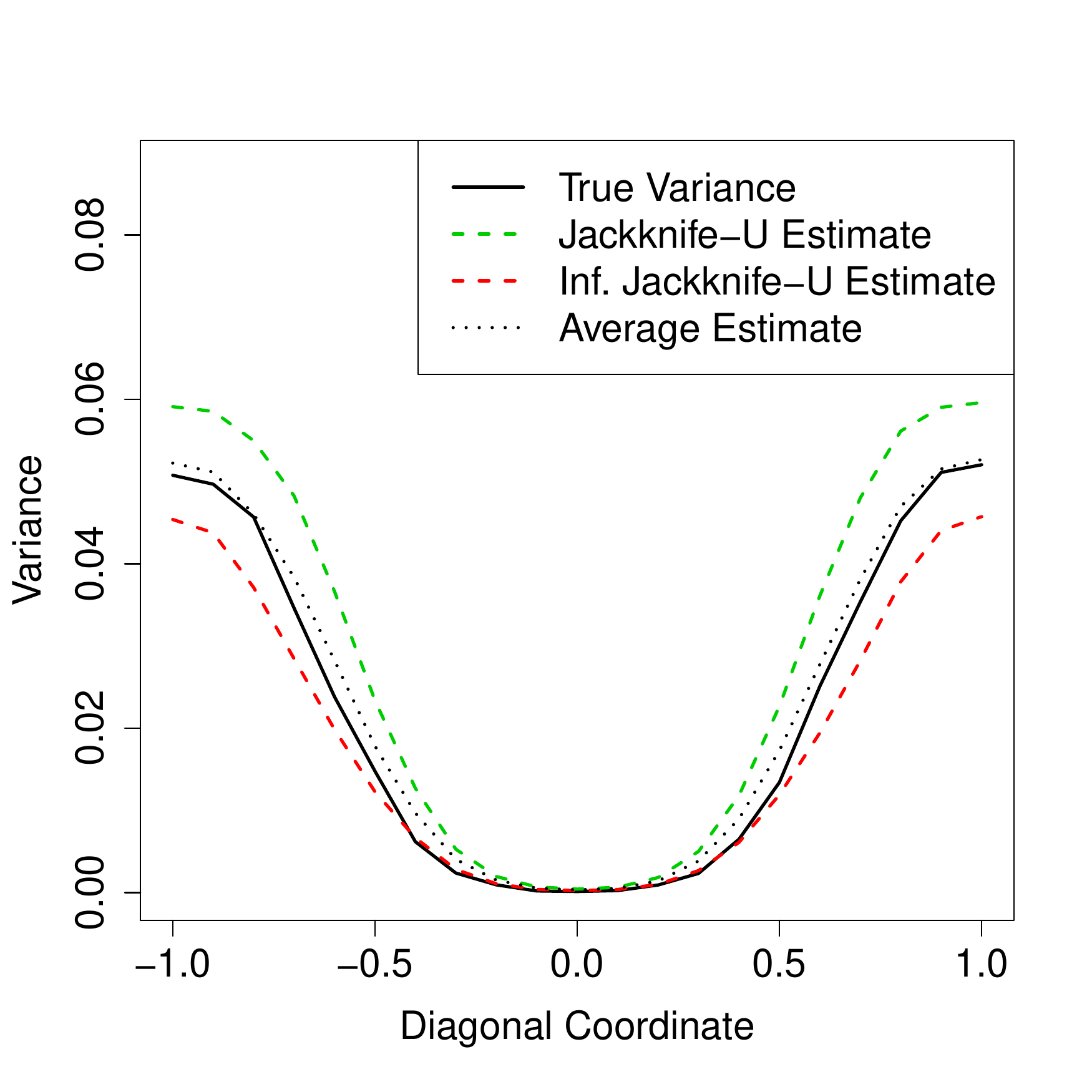}
\caption{Bagged tree sampling bias.}
\label{fig:tree_bias}
\end{subfigure}
\hspace{0.05\columnwidth}
\begin{subfigure}{0.45\columnwidth}
\includegraphics[width = \columnwidth]{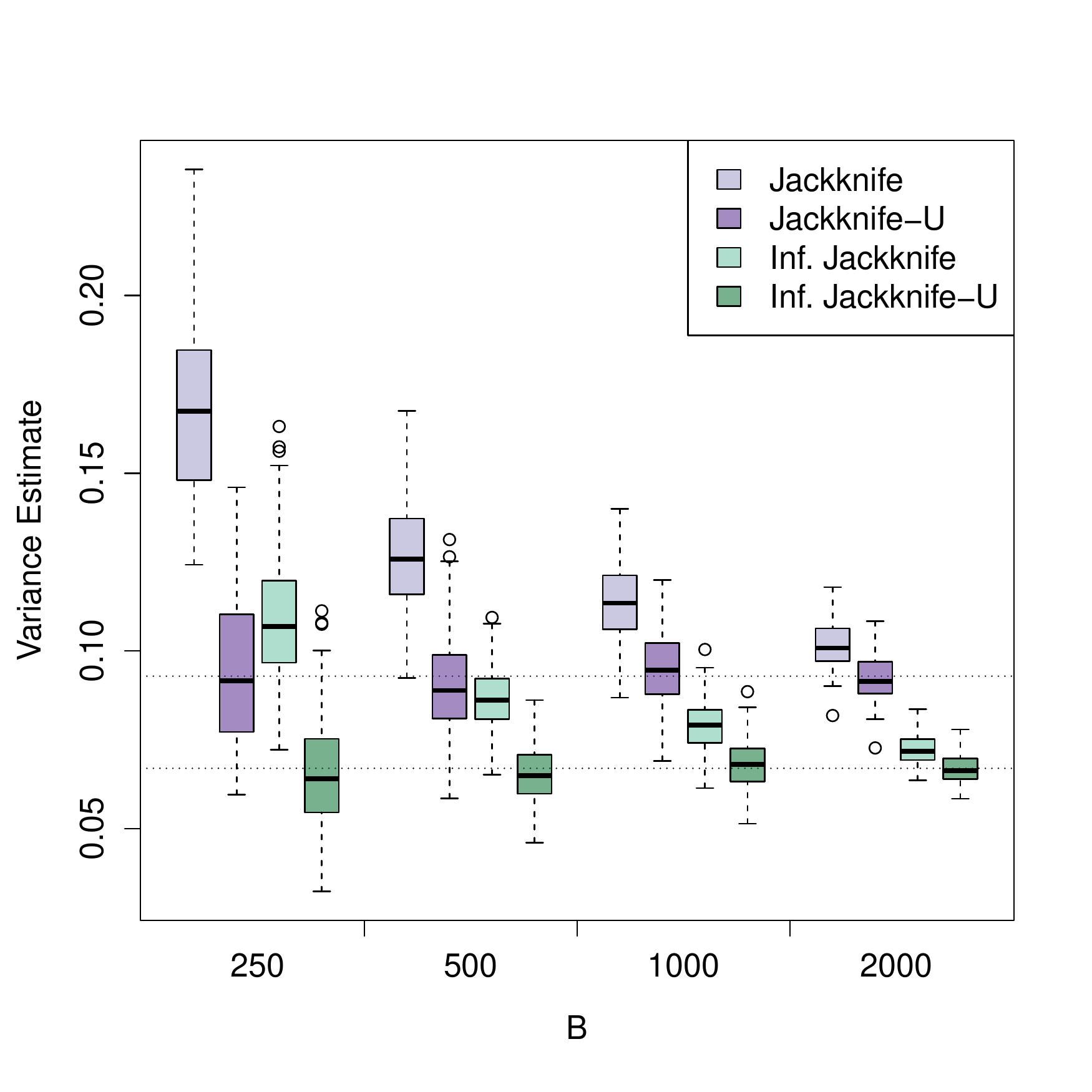}
\caption{Prostate cancer example.}
\label{fig:lpsa}
\end{subfigure}
\caption{Sampling bias of the jackknife and IJ rules. In the left panel, we compare the expected values of the jackknife and IJ estimators as well as their mean with the true variance of a bagged tree. In this example, the features take values in $(x_1, \, x_2) \in [-1, \, 1]^2$; we depict variance estimates along the diagonal $x_1 = x_2$. The prostate cancer plot can be interpreted in the same way as Figure \ref{fig:chol}, except that the we now indicate the weighted means of the J-U and IJ-U estimators separately.}
\label{fig:varbias}
\end{figure}

To test this idea, we used both the jackknife and IJ methods to estimate the variance of a bagged tree trained on a sample of size $n = 25$. (See Appendix \ref{sec:details} for details.) Since the sample size is so small, both the jackknife and IJ estimators exhibit some bias as seen in Figure \ref{fig:tree_bias}. However, the mean of the two estimators is nearly unbiased for the true variance of the bagged tree. (It appears that this mean has a very slight upward bias, just as we would expect from \eqref{eq:MEAN_BIAS}.)

This issue can arise in real datasets too. When training bagged forward stepwise regression on a prostate cancer dataset discussed by \citet{hastie2009elements}, the jackknife and IJ methods give fairly different estimates of variance: the jackknife estimator converged to 0.093, while the IJ estimator stabilized at 0.067 (Figure \ref{fig:lpsa}). Based on the discussion in this section, it appears that $(0.093 + 0.067)/2 = 0.08$ should be considered a more unbiased estimate of variance than either of the two numbers on their own. 

\begin{table}[t]
\begin{center}
\begin{tabular}{r|rrr||r|ccc|}
 Function & $n$ & $p$ & $B$ & ERR & $\hVIJU$ & $\hVJU$ & $\frac{1}{2}(\hVIJU + \hVJU)$ \\ 
  \hline
 &  &  &  & Bias & $ -0.15 \, (\pm 0.03) $ & $ 0.14 \, (\pm 0.02) $ & $ \mathbf{-0.01 \, (\pm 0.02)} $ \\ 
  Cosine & 50 & 2 & 200 & Var & $ \mathbf{0.08 \, (\pm 0.02)} $ & $ 0.41 \, (\pm 0.13) $ & $ 0.2 \, (\pm 0.06) $ \\ 
   &  &  &  & MSE & $ \mathbf{0.11 \, (\pm 0.03)} $ & $ 0.43 \, (\pm 0.13) $ & $ 0.2 \, (\pm 0.06) $ \\ 
   \hline
 &  &  &  & Bias & $ -0.05 \, (\pm 0.01) $ & $ 0.07 \, (\pm 0.01) $ & $ \mathbf{0.01 \, (\pm 0.01)} $ \\ 
  Cosine & 200 & 2 & 500 & Var & $ \mathbf{0.02 \, (\pm 0)} $ & $ 0.07 \, (\pm 0.01) $ & $ 0.04 \, (\pm 0.01) $ \\ 
   &  &  &  & MSE & $ \mathbf{0.02 \, (\pm 0)} $ & $ 0.07 \, (\pm 0.01) $ & $ 0.04 \, (\pm 0.01) $ \\ 
\hline
 &  &  &  & Bias & $ -0.3 \, (\pm 0.03) $ & $ 0.37 \, (\pm 0.04) $ & $ \mathbf{0.03 \, (\pm 0.03)} $ \\ 
  Noisy XOR & 50 & 50 & 200 & Var & $ \mathbf{0.48 \, (\pm 0.03)} $ & $ 1.82 \, (\pm 0.12) $ & $ 0.89 \, (\pm 0.05) $ \\ 
   &  &  &  & MSE & $ \mathbf{0.58 \, (\pm 0.03)} $ & $ 1.96 \, (\pm 0.13) $ & $ 0.89 \, (\pm 0.05) $ \\ 
   \hline
 &  &  &  & Bias & $ \mathbf{-0.08 \, (\pm 0.02)} $ & $ 0.24 \, (\pm 0.03) $ & $ \mathbf{0.08 \, (\pm 0.02)} $ \\ 
  Noisy XOR & 200 & 50 & 500 & Var & $ \mathbf{0.26 \, (\pm 0.02)} $ & $ 0.77 \, (\pm 0.04) $ & $ 0.4 \, (\pm 0.02) $ \\ 
   &  &  &  & MSE & $ \mathbf{0.27 \, (\pm 0.01)} $ & $ 0.83 \, (\pm 0.04) $ & $ 0.41 \, (\pm 0.02) $ \\ 
   \hline
 &  &  &  & Bias & $ \mathbf{-0.23 \, (\pm 0.04)} $ & $ 0.65 \, (\pm 0.05) $ & $ \mathbf{0.21 \, (\pm 0.04)} $ \\ 
  Noisy AND & 50 & 500 & 200 & Var & $ \mathbf{1.15 \, (\pm 0.05)} $ & $ 4.23 \, (\pm 0.18) $ & $ 2.05 \, (\pm 0.09) $ \\ 
   &  &  &  & MSE & $ \mathbf{1.21 \, (\pm 0.06)} $ & $ 4.64 \, (\pm 0.21) $ & $ 2.09 \, (\pm 0.09) $ \\ 
   \hline
 &  &  &  & Bias & $ \mathbf{-0.04 \, (\pm 0.04)} $ & $ 0.32 \, (\pm 0.04) $ & $ 0.14 \, (\pm 0.03) $ \\ 
  Noisy AND & 200 & 500 & 500 & Var & $ \mathbf{0.55 \, (\pm 0.07)} $ & $ 1.71 \, (\pm 0.22) $ & $ 0.85 \, (\pm 0.11) $ \\ 
   &  &  &  & MSE & $ \mathbf{0.57 \, (\pm 0.08)} $ & $ 1.82 \, (\pm 0.24) $ & $ 0.88 \, (\pm 0.11) $ \\ 
   \hline
   &  &  &  & Bias & $ -0.11 \, (\pm 0.02) $ & $ 0.23 \, (\pm 0.05) $ & $ \mathbf{0.06 \, (\pm 0.03)} $ \\ 
  Auto & 314 & 7 & 1000 & Var & $ \mathbf{0.13 \, (\pm 0.04)} $ & $ 0.49 \, (\pm 0.19) $ & $ 0.27 \, (\pm 0.1) $ \\ 
   &  &  &  & MSE & $ \mathbf{0.15 \, (\pm 0.04)} $ & $ 0.58 \, (\pm 0.24) $ & $ 0.29 \, (\pm 0.11) $ \\ 
\hline
  \end{tabular}
\caption{Simulation study. We evaluate the mean bias, variance, and MSE of different variance estimates $\hV$ for random forests.  Here, $n$ is the number of test examples used, $p$ is the number of features, and $B$ is the number of trees grown; the numbers in parentheses are 95\% confidence errors from sampling. The best methods for each evaluation metric are highlighted in bold. The data-generating functions are described in Appendix \ref{sec:details}.}
\label{tab:sim}
\end{center}
\end{table}

In the more extensive simulations presented in Table \ref{tab:sim}, averaging $\hVIJU$ and $\hVJU$ is in general less biased than either of the original estimators (although the ``Noisy AND'' experiment seems to provide an exception to this rule, suggesting that most of the bias of $\hVJU$ for this function is due to higher-order interactions). However, $\hVIJU$ has systematically lower variance, which allows it to win in terms of overall mean squared error. Thus, if unbiasedness is important, averaging $\hVIJU$ and $\hVJU$ seems like a promising idea, but $\hVIJU$ appears to be the better rule in terms of raw MSE minimization.

Finally, we emphasize that this relative bias result relies on the heuristic relationship \eqref{eq:IJ_BIAS}. While this approximation does not seem problematic for the first-order analysis presented in Section \ref{sec:varvar}, we may be concerned that the plug-in argument from Appendix \ref{sec:varvartheory} used to justify it may not give us correct second- and higher-order terms. Thus, although our simulation results seem promising, developing a formal and general understanding of the relative biases of $\hVIJinf$ and $\hVJinf$ remains an open topic for follow-up research.

\section{Conclusion}

In this paper, we studied the jackknife-after-bootstrap and infinitesimal jackknife (IJ) methods \citep{efron1992jackknife, efron2012model} for estimating the variance of bagged predictors. We demonstrated that both estimators suffer from considerable Monte Carlo bias, and we proposed bias-corrected versions of the methods that appear to work well in practice. We also provided a simple formula for the sampling variance of the IJ estimator, and showed that from a sampling bias point of view the arithmetic mean of the jackknife and IJ estimators is often preferable to either of the original methods. Finally, we applied these methods in numerous experiments, including some random forest examples, and showed how they can be used to gain valuable insights in realistic problems.

\section*{Acknowledgments}

The authors are grateful for helpful suggestions from the action editor and three anonymous referees. S.W. is supported by a B.C. and E.J. Eaves Stanford Graduate Fellowship.

\begin{appendix}

\section{The Effect of Monte Carlo Noise on the Jackknife Estimator}
\label{sec:mc_JAB}

In this section, we derive expressions for the finite-$B$ Monte Carlo bias and variance of the jackknife-after-bootstrap estimate of variance. Recall from \eqref{eq:JACK} that
$$\hVJ = \frac{n-1}{n} \sum_{i = 1}^n \hDelta_i^2, \text{ where } \hDelta_i = \frac{\sum_{\{b \, : \, N^*_{bi} = 0\}} t_b^*}{\left\lvert\{N^*_{bi} = 0\}\right\rvert} - \frac{\sum_b t_b^*}{B} $$
and $N_{bi}^*$ indicates the number of times the $i^{th}$ observation appears in the bootstrap sample $b$. If $\hDelta_i$ is not defined because $N^*_{bi} = 0$ for either all or none of the $b = 1, \, ..., \, B$, then just set $\hDelta_i = 0$.

Now $\hVJ$ is the sum of squares of noisy quantities, and so $\hVJ$ will be biased upwards. Specifically,
$$ \EE_*\left[\hVJ\right] - \hVJinf = \frac{n - 1}{n}\sum_{i = 1}^n \Var_*\left[\hDelta_i\right], $$
where $\hVJinf$ is the jackknife estimate computed with $B = \infty$ bootstrap replicates. For convenience, let
$$ B_i = \left\lvert\{b : N_{bi} = 0\}\right\rvert, $$
and recall that
$$ \Var_*\left[\hDelta_i\right] = \EE_*\left[\Var_*\left[\hDelta_i|B_i\right]\right] + \Var_*\left[\EE_*\left[\hDelta_i|B_i\right]\right]. $$
For all $B_i \neq 0$ or $B$, the conditional expectation is
$$\EE_*[\hDelta_i|B_i] = \left(1 - \frac{B_i - \EE\left[B_i\right]}{B}\right) \Delta_i, \where \Delta_i = \EE_*\left[t^*_b|N^*_{bi} = 0\right] - \EE_*\left[t^*_b\right]; $$
$\EE_*[\hDelta_i|B_i] = 0$ in the degenerate cases with $B_i \in \{0, \, B\}$. Thus,
$$\Var_*\left[\EE_*\left[\hDelta_i|B_i\right]\right] = \oo\left(\Delta_i^{2} \big/ B\right), $$
and so
$$ \Var_*\left[\hDelta_i\right] = \EE_*\left[\Var_*\left[\hDelta_i|B_i\right]\right] + \oo\left(\Delta_i^{2} \big/ B\right). $$
Meanwhile, for $B_i \notin \{0, \, B\}$,
\begin{align*}
\Var_*\left[\hDelta_i|B_i\right]
&= \frac{1}{B^2}\left(\left(\frac{B}{B_i} - 1\right)^2 B_i \vn_i + (B - B_i)\tvn_i\right) \\
&= \frac{1}{B}\left(\frac{(B - B_i)^2}{BB_i} \, \vn_i + \frac{B - B_i}{B} \tvn_i\right)
\end{align*}
where
\begin{align*}
\vn_i = \Var_*\left[t^*_b|N^*_{bi} = 0\right] \text{ and }
\tvn_i = \Var_*\left[t^*_b|N^*_{bi} \neq 0\right].
\end{align*}
Thus,
\begin{align*}
\Var_*\left[\hDelta_i\right] &= \frac{1}{B}\left(\EE_*\left[\frac{(B - B_i)^2}{BB_i} 1_i \right] \, \vn_i + \EE_*\left[\frac{B - B_i}{B} 1_i \right] \tvn_i\right)
+ \oo\left(\Delta_i^{2} \big/ B\right),
\end{align*}
where $1_i = 1(\{B_i \notin \{0, \, B\}\})$.

As $n$ and $B$ get large, $B_i$ converges in law to a Gaussian random variable
$$ \frac{B_i - B e^{-1}}{\sqrt{B}} \Rightarrow \left(0, \, e^{-1} (1 - e^{-1})\right) $$
and the above expressions are uniformly integrable. We can verify that
$$ \EE_*\left[\frac{(B - B_i)^2}{BB_i} 1_i\right] = e - 2 + e^{-1} + \oo\left(\frac{1}{B}\right), $$
and
$$ \EE_*\left[\frac{B - B_i}{B} 1_i\right] = \frac{e - 1}{e} + \oo\left(\left(1 - e^{-1}\right)^B\right). $$
Finally, this lets us conclude that
\begin{align*}
\EE_*\left[\hVJ\right] - \hVJinf &= \frac{1}{B}\frac{n-1}{n}\sum_{i = 1}^n \left(\left(\frac{(e - 1)^2}{e}\right)\vn_i + \left(\frac{e - 1}{e}\right) \tvn_i\right)
+ \oo\left(\frac{1}{B} + \frac{n}{B^2}\right),
\end{align*}
where the error term depends on $\vn_i$, $\tvn_i$, and $\hVJinf = (n-1)/n \, \sum_{i = 1}^n \Delta_i^2$. 

We now address Monte Carlo variance. By the central limit theorem, $\hDelta_i$ converges to a Gaussian random variable as $B$ gets large. Thus, the asymptotic Monte Carlo variance of $\hDelta_i^2$ is approximately $2\Var_*[\hDelta_i]^2 + 4\EE_*[\hDelta_i]^2\Var_*[\hDelta_i]$, and so
\begin{align*}
\Var_*\left[\hVJ\right] &\approx 2 \, \left(\frac{1}{B}\frac{n-1}{n}\right)^2 \sum_{i = 1}^n \left(\left(\frac{(e - 1)^2}{e}\right)\vn_i + \left(\frac{e - 1}{e}\right) \tvn_i\right)^2 \\
& \ \ \ \ \ \ \ \ + 4 \, \frac{1}{B}\frac{n-1}{n} \sum_{i = 1}^n \Delta_i^2 \left(\left(\frac{(e - 1)^2}{e}\right)\vn_i + \left(\frac{e - 1}{e}\right) \tvn_i\right).
\end{align*}
In practice, the terms $\vn_i$ and $\tvn_i$ can be well approximated by $\vboot = \Var_*[t_b^*]$, namely the bootstrap estimate of variance for the base learner. (Note that $\vn_i$, $\tvn_i$, and $\vboot$ can always be inspected on a random forest, so this assumption can be checked in applications.) This lets us considerably simplify our expressions for Monte Carlo bias and variance:
\begin{align*}
&\EE_*\left[\hVJ\right] - \hVJinf \approx \frac{n}{B} \, (e -1)\vboot, \text{ and }\\
&\Var_*\left[\hVJ\right] \approx 2 \, \frac{n}{B^2} \, (e-1)^2 \hv^2 + 4 \, \frac{1}{B} \, (e-1) \, \hVJinf \, \vboot.
\end{align*}

\section{The IJ estimator and H{\'a}jek projections}
\label{sec:varvartheory}

Up to \eqref{eq:boothaj}, the derivation below is an alternate presentation of the argument made by \citet{efron2012model} in the proof of his Theorem 1.
To establish a connection between the IJ estimate of variance for bagged estimators and the theory of H\'ajek projections, it is useful to consider $\hthBAG$ as a functional over distributions. Let $G$ be a probability distribution, and let $T$ be a functional over distributions with the following property:
\begin{equation}
\label{eq:averaging}
 T(G) = \EE_G[\tau(Y_1, \, ..., \, Y_n)] \text{ for some function } \tau,
\end{equation}
where the $Y_1, \, ..., \, Y_n$ are drawn independently from $G$. We call functionals $T$ satisfying \eqref{eq:averaging} \emph{averaging}. Clearly, $\hthBAG$ can be expressed as an averaging functional applied to the empirical distribution $\hF$ defined by the observations $Z_1, \, ..., \, Z_n$.

Suppose that we have an averaging functional $T$,  a sample $Z_1, \, ..., \, Z_n$ forming an empirical distribution $\hF$, and want to study the variance of $T(\hF)$. The infinitesimal jackknife estimate for the variance of $\hT$ is given by
$$ \hV = \sum_{i = 1}^n \left(\frac{1}{n} \frac{\partial}{\partial \varepsilon} T\left(\hF_i(\varepsilon)\right)\right)^2, $$
where $\hF_i(\varepsilon)$ is the discrete distribution that places weight $1/n + (n-1)/n \cdot \varepsilon$ at $Z_i$ and weight $1/n - \varepsilon/n$ at all the other $Z_j$.

We can transform samples from $\hF$ into samples from $\hF_i(\varepsilon)$ by the following method. Let $Z^*_1, \, ..., \, Z^*_n$ be a sample from $\hF$. Go through the whole sample and, independently for each $j$, take $Z_j^*$ and with probability $\varepsilon$ replace it with $Z_i$. The sample can now be considered a sample from $\hF_i(\varepsilon)$.

When $\varepsilon \rightarrow 0$, the probability of replacing two of the $Z_i^*$ with this procedure becomes negligible, and we can equivalently transform our sample into a sample from $\hF_i(\varepsilon)$ by transforming a single random element from $\{Z_j^*\}$ into $Z_i$ with probability $n \, \varepsilon$. Without loss of generality this element is the first one, and so we conclude that
\begin{align*}
\lim_{\varepsilon \rightarrow 0} \frac{1}{\varepsilon} &\left(\pEEsub{\hF_i(\varepsilon)}{\tau(Z_1^*, \, ..., \, Z_n^*)} - 
\pEEsub{\hF}{\tau(Z_1^*, \, ..., \, Z_n^*)}\right)\\
& = n \, \left(\pEEsub{\hF}{\tau(Z_1^*, \, ..., \, Z_n^*) \big | Z_1^* = Z_i} - \pEEsub{\hF}{\tau(Z_1^*, \, ..., \, Z_n^*)}\right),
\end{align*}
where $\tau$ defines $T$ through \eqref{eq:averaging}. Thus,
$$ \frac{1}{n} \frac{\partial}{\partial \varepsilon} T(\hF_i(\varepsilon)) =\pEEsub{\hF}{T \big | Z_1^* = Z_i} - \pEEsub{\hF}{T}, $$
and so
\begin{align}
\label{eq:boothaj}
\hV
&= \sum_{i = 1}^n \left(\pEEsub{\hF}{T \big | Z_1^* = Z_i} - \pEEsub{\hF}{T}\right)^2 \\
\label{eq:plugin}
&\approx  \sum_{i = 1}^n \left(\pEEsub{F}{T \big | Z_1^* = Z_i} - \pEEsub{F}{T}\right)^2,
\end{align}
where on the last line we only replaced the empirical approximation $\hF$ with its true value $F$. In the case of bagging, this last expression is equivalent to \eqref{eq:approxhajek}.

A crucial step in the above argument is the plug-in approximation \eqref{eq:plugin}. If $T$ is just a sum, then the error of \eqref{eq:plugin} is within $\oo(1/n)$; presumably, similar statements hold whenever $T$ is sufficiently well-behaved. That being said, it is possible to construct counter-examples where \eqref{eq:plugin} fails; a simple such example is when $T$ counts the number of times $Z_1^*$ is matched in the rest of the training data. Establishing general conditions under which \eqref{eq:plugin} holds is an interesting topic for further research.

\section{Description of Experiments}
\label{sec:details}

This section provides a more detailed description of the experiments reported in this paper.

\subsubsection*{Auto MPG Example (Figre \ref{fig:auto})}

The Auto MPG dataset, available from the UCI Machine Learning Repository \citep{UCI}, is a regression task with 7 features. After discarding examples with missing entries, the dataset had 392 rows, which we divided into a test set of size 78 and a train set of size 314. We estimated the variance of the random forest predictions using the $(\hVJU + \hVIJU)/2$ estimator advocated in Section \ref{sec:bias}, with $B = 10,000$ bootstrap replicates.

\subsubsection*{Bagged Tree Simulation (Figure \ref{fig:example})}
\begin{figure}[t]
\centering
\includegraphics[width = 0.4\columnwidth]{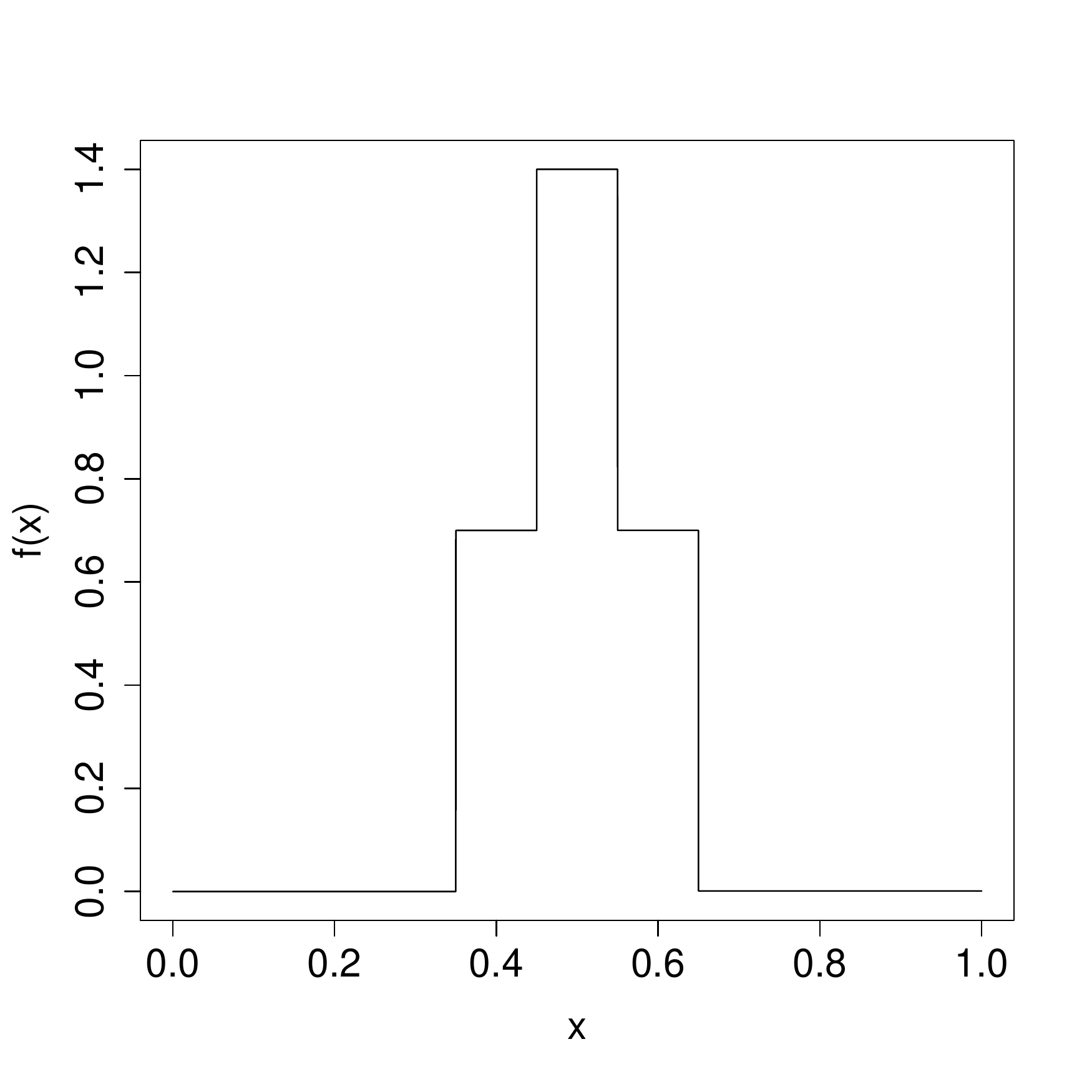}
\caption{Underlying model for the bagged tree example from Figure \ref{fig:example}.}
\label{fig:tree_model}
\end{figure}

The data for this simulation was drawn from a model $y_i = f(x_i) + \varepsilon_i$, where $x_i \sim U([0, 1])$, $\varepsilon_i \sim \nn(0, 1/2^2)$, and $f(x)$ is the step function shown in Figure \ref{fig:tree_model}. We modeled the data using 5-leaf regression trees generated using the \texttt{R} package \texttt{tree} \citep{ripley2002modern}; for bagging, we used $B = 10,000$ bootstrap replicates. The reported data is compiled over $1,000$ simulation runs with $n = 500$ data points each.

\subsubsection*{Cholesterol Example (Figures \ref{fig:chol}, \ref{fig:chol_ratio}, and \ref{fig:varvar})}
For the cholesterol dataset \citep{efron1991compliance}, we closely follow the methodology of \citet{efron2012model}; see his paper for details. The dataset has $n = 164$ subjects and only one predictor.

\subsubsection*{E-mail Spam Example (Figures \ref{fig:spam} and \ref{fig:spam_cmp})}

The e-mail spam dataset \citep[spambase,][]{UCI} is a classification problem with $n = 4,601$ e-mails and $p = 57$ features; the goal is to discern spam from non-spam. We divided the data into train and test sets of size 3,065 and 1,536 respectively. Each of the random forests described in Section \ref{sec:email} was fit on the train set using the \texttt{R} package \texttt{randomForest} \citep{liaw2002classification} with $B = 40,000$ bootstrap replicates.

\subsubsection*{California Housing Example (Figure \ref{fig:housing})}

The California housing dataset \citep[described in][and available from StatLib]{hastie2009elements} contains aggregated data from $n = 20,460$ neighborhoods. There are $p = 8$ features; the response is the median house value. We fit random forests on this data using the \texttt{R} package \texttt{randomForest} \citep{liaw2002classification} with $B = 1,000$ bootstrap replicates. 

\subsubsection*{Bagged Tree Simulation \#2 (Figure \ref{fig:tree_bias})}

We drew $n = 25$ points from a model where the $x_i$ are uniformly distributed over a square, i.e., $x_i \sim U([-1, 1]^2)$; the $y_i$ are deterministically given by $y_i = 1(\{||x_i||_2 \geq 1\})$. We fit this data using the \texttt{R} package \texttt{tree} \citep{ripley2002modern}. The bagged predictors were generated using $B = 1,000$ bootstrap replicates. The reported results are based on 2,000 simulation runs.

\subsubsection*{Prostate Cancer Example (Figure \ref{fig:lpsa})}

The prostate cancer data \citep[published by][]{stamey1989prostate} is described in Section 1 of \citet{hastie2009elements}. We used forward stepwise regression as implemented by the \texttt{R} function \texttt{step} as our base learner. This dataset has $n = 97$ subjects and $8$ available predictor variables. In figure \ref{fig:lpsa}, we display standard errors for the predicted response of a patient whose features match those of patient $\#41$ in the dataset.

\subsubsection*{Simulations for Table \ref{tab:sim}}

The data generation functions used in Table \ref{tab:sim} are defined as follows. The $X_i$ for $i = 1, \, ..., \, p$ are all generated as independent $U([0,\,1])$ random variables, and $\varepsilon \sim \nn(0, 1)$.
\begin{itemize}
\item Cosine: $Y = 3 \cdot \cos\left(\pi \cdot (X_1 + X_2)\right)$, with $p = 2$.
\item Noisy XOR: Treating $\texttt{XOR}$ as a function with a $0/1$ return-value,
$$Y = 5 \cdot \left[\texttt{XOR}\left(X_1 > 0.6, \, X_2 > 0.6\right) + \texttt{XOR}\left(X_3 > 0.6, \, X_4 > 0.6\right)\right] + \varepsilon $$
and $p = 50$.
\item Noisy AND: With analogous notation,
$$Y = 10 \cdot \texttt{AND}\left(X_1 > 0.3, \, X_2 > 0.3, \, X_3 > 0.3, \, X_4 > 0.3 \right) + \varepsilon $$
and $p = 500$.
\item Auto: This example is based on a parametric bootstrap built on the same dataset as used in Figure \ref{fig:auto}. We first fit a random forest to the training set, and evaluated the MSE $\hsigma^2$ on the test set. We then generated new training sets by replacing the labels $Y_i$ from the original training set with $\hY_i + \hsigma \varepsilon$, where $\hY_i$ is the original random forest prediction at the $i^{th}$ training example and $\varepsilon$ is fresh residual noise.
\end{itemize}
During the simulation, we first generated a random test set of size 50 (except for the auto example, where we just used the original test set of size 78). Then, while keeping the test set fixed, we generated 100 training sets and produced variance estimates $\hV$ at each test point. Table \ref{tab:sim} reports average performance over the test set.

\end{appendix}

\bibliography{./references}

\end{document}